\documentclass[twoside,11pt]{article}

\usepackage{jmlr2e}

\usepackage{amsmath}
\usepackage{multirow}
\usepackage{booktabs}
\usepackage[english]{babel}
\usepackage{url}
\usepackage{siunitx}

\jmlrheading{17}{2016}{1-32}{10/15; Revised 4/16}{4/16}{Jure \v{Z}bontar and Yann LeCun}

\ShortHeadings{Stereo by Training a Network to Compare Patches}{\v{Z}bontar and LeCun}
\firstpageno{1}

\begin{document}

\title{Stereo Matching by Training a Convolutional Neural Network to Compare Image Patches}

\author{\name Jure \v{Z}bontar\thanks{
    Jure \v{Z}bontar is also with the
    \addr Courant Institute of Mathematical Sciences,
    New York University,
    715 Broadway, New York, NY 10003, USA.
} \email jure.zbontar@fri.uni-lj.si \\
\addr Faculty of Computer and Information Science\\
University of Ljubljana\\
Ve\v{c}na pot 113, SI-1001 Ljubljana, Slovenia
\AND
\name Yann LeCun\thanks{
    Yann LeCun is also with
    \addr Facebook AI Research,
    770 Broadway, New York, NY 10003, USA.
} \email yann@cs.nyu.edu \\
\addr Courant Institute of Mathematical Sciences\\
New York University\\
715 Broadway, New York, NY 10003, USA}

\editor{Zhuowen Tu}

\maketitle

\begin{abstract}%
We present a method for extracting depth information from a rectified image
pair. Our approach focuses on the first stage of many stereo algorithms: the
matching cost computation. We approach the problem by learning a similarity
measure on small image patches using a convolutional neural network. Training
is carried out in a supervised manner by constructing a binary classification
data set with examples of similar and dissimilar pairs of patches. We examine
two network architectures for this task: one tuned for speed, the other for
accuracy. The output of the convolutional neural network is used to initialize
the stereo matching cost. A series of post-processing steps follow: cross-based
cost aggregation, semiglobal matching, a left-right consistency check, subpixel
enhancement, a median filter, and a bilateral filter. We evaluate our method on
the KITTI 2012, KITTI 2015, and Middlebury stereo data sets and show that it
outperforms other approaches on all three data sets.

\end{abstract}


\section{Introduction}
\begin{figure*}[t]
\begin{center}
\includegraphics[scale=0.95]{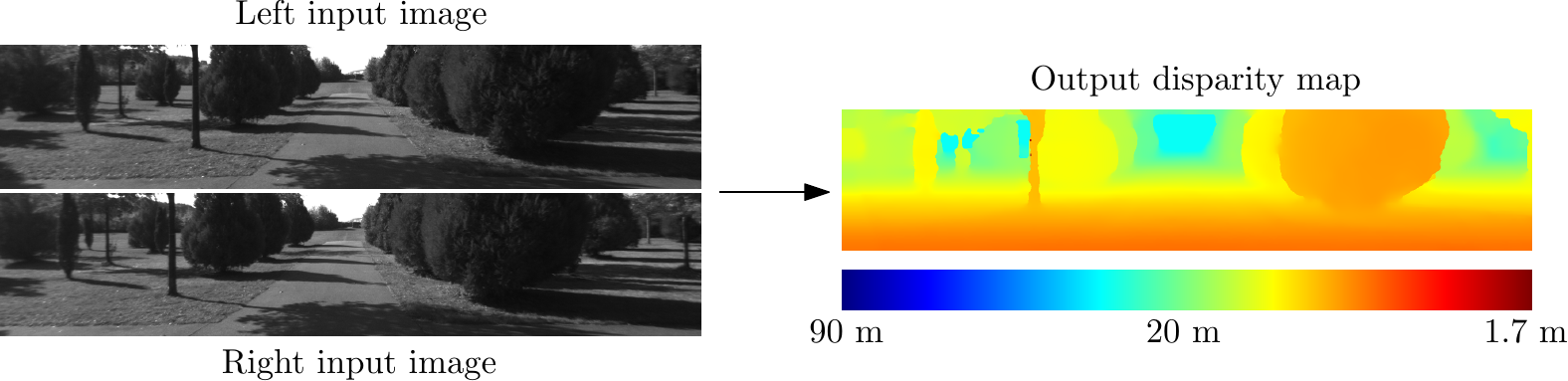}
\end{center}
\caption{The input is a pair of images from the left and right camera. The two
input images differ mostly in horizontal locations of objects (other
differences are caused by reflections, occlusions, and perspective distortions).
Note that objects closer to the camera have larger disparities than objects
farther away. The output is a dense disparity map shown on the right, with
warmer colors representing larger values of disparity (and smaller values of
depth).} 

\label{fig:input_output}
\end{figure*}

Consider the following problem: given two images taken by cameras at
different horizontal positions, we wish to compute the disparity $d$ for each
pixel in the left image. Disparity refers to the difference in horizontal
location of an object in the left and right image---an object at position $(x,
y)$ in the left image appears at position $(x - d, y)$ in the right image. If
we know the disparity of an object we can compute its depth $z$ using the
following relation:
\begin{equation*}
z = \frac{f B}{d},
\end{equation*}
where $f$ is the focal length of the camera and $B$ is the distance between the
camera centers. Figure~\ref{fig:input_output} depicts the input to and the
output from our method. 

The described problem of stereo matching is important in many fields such as
autonomous driving, robotics, intermediate view generation, and 3D scene
reconstruction. According to the taxonomy of \citet{scharstein2002taxonomy}, a
typical stereo algorithm consists of four steps: matching cost computation,
cost aggregation, optimization, and disparity refinement. Following
\citet{hirschmuller2009evaluation} we refer to the first two steps as computing
the matching cost and the last two steps as the stereo method. 
The focus of this work is on computing a good matching cost. 

We propose training a convolutional neural network \citep{lecun1998gradient} on
pairs of small image patches where the true disparity is known (for example,
obtained by LIDAR or structured light). The output of the network is used to
initialize the matching cost. We proceed with a number of post-processing steps
that are not novel, but are necessary to achieve good results. Matching costs
are combined between neighboring pixels with similar image intensities using
cross-based cost aggregation. Smoothness constraints are enforced by semiglobal
matching and a left-right consistency check is used to detect and eliminate
errors in occluded regions. We perform subpixel enhancement and apply a median
filter and a bilateral filter to obtain the final disparity map. 

The contributions of this paper are
\begin{itemize} 
\setlength{\itemsep}{1pt}
\item a description of two architectures based on convolutional neural
networks for computing the stereo matching cost;

\item a method, accompanied by its source code, with the lowest error rate on
the KITTI 2012, KITTI 2015, and Middlebury stereo data sets; and

\item experiments analyzing the importance of data set size, the error rate
compared with other methods, and the trade-off between accuracy and runtime for
different settings of the hyperparameters.

\end{itemize}

This paper extends our previous work \citep{Zbontar_2015_CVPR} by including
a description of a new architecture, results on two new data sets,
lower error rates, and more thorough experiments.

\section{Related Work}

Before the introduction of large stereo data sets like KITTI and Middlebury,
relatively few stereo algorithms used ground truth information to learn
parameters of their models; in this section, we review the ones that did. For a
general overview of stereo algorithms see \citet{scharstein2002taxonomy}.

\citet{kong2004method} used the sum of squared distances to compute an initial
matching cost. They then trained a model to predict the probability distribution
over three classes: the initial disparity is correct, the initial disparity is
incorrect due to fattening of a foreground object, and the initial disparity is
incorrect due to other reasons. The predicted probabilities were used to adjust
the initial matching cost. \citet{kong2006stereo} later extend their work by
combining predictions obtained by computing normalized cross-correlation over
different window sizes and centers.  \citet{peris2012towards} initialized the
matching cost with AD-Census \citep{mei2011building}, and used multiclass
linear discriminant analysis to learn a mapping from the computed matching cost
to the final disparity.

Ground-truth data was also used to learn parameters of probabilistic graphical
models. \citet{zhang2007estimating} used an alternative optimization algorithm
to estimate optimal values of Markov random field hyperparameters.
\citet{scharstein2007learning} constructed a new data set of 30 stereo pairs and
used it to learn parameters of a conditional random field.
\citet{li2008learning} presented a conditional random field model with a
non-parametric cost function and used a structured support vector machine to
learn the model parameters.

Recent work \citep{haeusler2013ensemble,spyropoulos2014learning} focused on
estimating the confidence of the computed matching cost.
\citet{haeusler2013ensemble} used a
random forest classifier to combine several confidence measures. Similarly,
\citet{spyropoulos2014learning} trained a random forest classifier to predict
the confidence of the matching cost and used the predictions as soft constraints
in a Markov random field to decrease the error of the stereo method.

A related problem to computing the matching cost is learning local image
descriptors \citep{brown2011discriminative, trzcinski2012learning,
simonyan2014learning, revaud2015deepmatching, paulin2015local, han2015matchnet,
zagoruyko2015learning}. The two problems share a common subtask: to measure the
similarity between image patches.  \citet{brown2011discriminative} introduced a
general framework for learning image descriptors and used Powell's method to
select good hyperparameters.  Several methods have been suggested for solving
the problem of learning local image descriptors, such as 
boosting \citep{trzcinski2012learning}, 
convex optimization \citep{simonyan2014learning}, 
hierarchical moving-quadrant similarity \citep{revaud2015deepmatching},
convolutional kernel networks \citep{paulin2015local}, 
and convolutional neural networks \citep{zagoruyko2015learning,han2015matchnet}. 
Works of \citet{zagoruyko2015learning} and \citet{han2015matchnet}, in
particular, are very similar to our own, differing mostly in the architecture of
the network; concretely, the inclusion of pooling and subsampling to account
for larger patch sizes and larger variation in viewpoint.

\section{Matching Cost}

A typical stereo algorithm begins by computing a matching cost at each
position $\mathbf{p}$ for all disparities $d$ under consideration. A
simple method for computing the matching cost is the sum of absolute
differences:
\begin{equation} C_{\text{SAD}}(\mathbf{p}, d) = \sum_{\mathbf{q} \in
\mathcal{N}_{\mathbf{p}}} |I^L(\mathbf{q}) - I^R(\mathbf{q}-\mathbf{d})|,
\label{eqn:C_sad}
\end{equation}
where $I^L(\mathbf{p})$ and $I^R(\mathbf{p})$ are image intensities at position
$\mathbf{p}$ in the left and right image and $\mathcal{N}_{\mathbf{p}}$ is the
set of locations within a fixed rectangular window centered at $\mathbf{p}$. 

We use bold lowercase letters $\mathbf{p}$ and $\mathbf{q}$ to denote image
locations. A bold lowercase $\mathbf{d}$ denotes the disparity $d$ cast to a
vector, that is, $\mathbf{d} = (d, 0)$.  We use \texttt{typewriter} font for
the names of hyperparameters. For example, we would use \texttt{patch\_size} to
denote the size of the neighbourhood area $\mathcal{N}_{\mathbf{p}}$.

Equation~\eqref{eqn:C_sad} can be interpreted as measuring the cost
associated with matching a patch from the left image, centered at position
$\mathbf{p}$, with a patch from the right image, centered at position
$\mathbf{p}-\mathbf{d}$. We want the cost to be low when the two patches are centered
around the image of the same 3D point, and high when they are not. 

Since examples of good and bad matches can be constructed from publicly
available data sets (for example, the KITTI and Middlebury stereo data sets), we
can attempt to solve the matching problem by a supervised learning approach.
Inspired by the successful application of convolutional neural networks to
vision problems, we used them to assess how well two small image patches match.

\subsection{Constructing the Data Set}
\label{sec:dataset}

We use ground truth disparity maps from either the KITTI or Middlebury stereo
data sets to construct a binary classification data set. At each image position
where the true disparity is known we extract one negative and one positive
training example. This ensures that the data set contains an equal number of
positive and negative examples. A positive example is a pair of patches, one
from the left and one from the right image, whose center pixels are the
images of the same 3D point, while a negative example is a pair of patches
where this is not the case. The following section describes the data set
construction step in detail.

Let
$
<\mathcal{P}_{n \times n}^L(\mathbf{p}), \mathcal{P}_{n \times n}^R
(\mathbf{q})>
$
denote a pair of patches, where $\mathcal{P}_{n \times n}^L(\mathbf{p})$ is an
$n \times n$ patch from the left image centered at position $\mathbf{p} = (x, y)$,
$\mathcal{P}_{n \times n}^R(\mathbf{q})$ is an $n \times n$ patch from the right image
centered at position $\mathbf{q}$, and $d$ denotes the correct disparity at position
$\mathbf{p}$. A negative example is obtained by setting the center of the right
patch to
\begin{equation*}
\mathbf{q} = (x - d + o_{\text{neg}}, y),
\end{equation*}
where $o_{\text{neg}}$ is chosen from either the interval
\([\texttt{dataset\_neg\_low}, \texttt{dataset\_neg\_high}]\) or, its origin
reflected counterpart, \([-\texttt{dataset\_neg\_high},
-\texttt{dataset\_neg\_low}]\). The random offset $o_{\text{neg}}$ ensures
that the resulting image patches are not centered around the same 3D point. 

A positive example is derived by setting
\begin{equation*}
\mathbf{q} = (x - d + o_{\text{pos}}, y),
\end{equation*}
where $o_{\text{pos}}$ is chosen randomly from the interval
\([-\texttt{dataset\_pos}, \texttt{dataset\_pos}]\).  The reason for including
$o_{\text{pos}}$, instead of setting it to zero, has to do with the stereo
method used later on. In particular, we found that cross-based cost aggregation
performs better when the network assigns low matching costs to good matches as
well as near matches. In our experiments, the hyperparameter \texttt{dataset\_pos}
was never larger than one pixel.

\subsection{Network Architectures}

We describe two network architectures for learning a similarity measure on
image patches. The first architecture is faster than the second, but produces
disparity maps that are slightly less accurate. In both cases, the input to the
network is a pair of small image patches and the output is a measure of
similarity between them. Both architectures contain a trainable feature
extractor that represents each image patch with a feature vector. The
similarity between patches is measured on the feature vectors instead of the
raw image intensity values. The fast architecture uses a fixed similarity
measure to compare the two feature vectors, while the accurate architecture
attempts to learn a good similarity measure on feature vectors. 

\subsubsection{Fast Architecture}

\begin{figure}[tb]
\begin{center}
\includegraphics{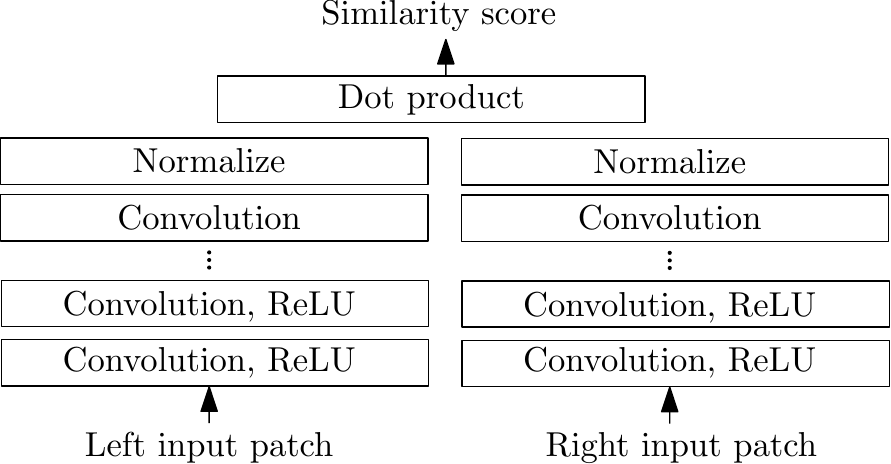}
\end{center}
\caption{The fast architecture is a siamese network. The two sub-networks
consist of a number of convolutional layers followed by rectified linear units
(abbreviated ``ReLU''). The similarity score is obtained by extracting a vector
from each of the two input patches and computing the cosine similarity between
them. In this diagram, as well as in our implementation, the cosine similarity
computation is split in two steps: normalization and dot product. This reduces
the running time because the normalization needs to be performed only once per
position (see Section~\ref{sec:matching_cost}).}

\label{fig:architecture_fast}
\end{figure}

The first architecture is a siamese network, that is, two shared-weight
sub-networks joined at the head \citep{bromley1993signature}.  The sub-networks
are composed of a number of convolutional layers with rectified linear units
following all but the last layer. Both sub-networks output a vector capturing
the properties of the input patch. The resulting two vectors are compared using
the cosine similarity measure to produce the final output of the network.
Figure~\ref{fig:architecture_fast} provides an overview of the architecture. 

The network is trained by minimizing a hinge loss. The loss is computed by
considering pairs of examples centered around the same image position where one
example belongs to the positive and one to the negative class. Let \( s_{+}
\) be the output of the network for the positive example, \( s_{-} \) be the
output of the network for the negative example, and let \(m\), the margin, be a
positive real number. The hinge loss for that pair of examples is defined as
$\max(0, m + s_{-} - s_{+})$. The loss is zero when the similarity of the
positive example is greater than the similarity of the negative example by at
least the margin \(m\). We set the margin to 0.2 in our experiments.

The hyperparameters of this architecture are 
the number of convolutional layers in each sub-network (\texttt{num\_conv\_layers}), 
the size of the convolution kernels (\texttt{conv\_kernel\_size}), 
the number of feature maps in each layer (\texttt{num\_conv\_feature\_maps}), 
and the size of the input patch (\texttt{input\_patch\_size}).

\subsubsection{Accurate Architecture}

\begin{figure}[tb] 
\begin{center}
\includegraphics{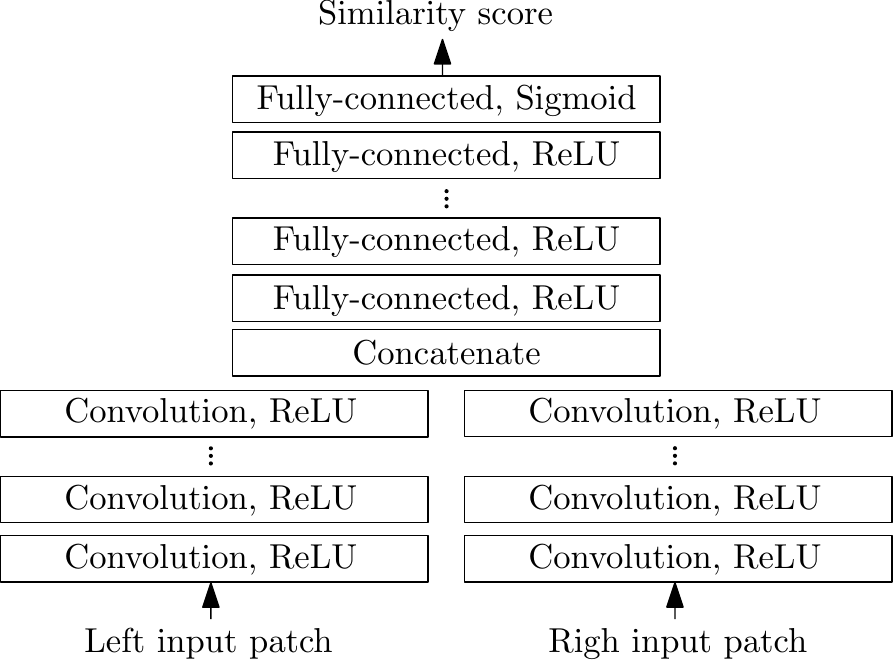} 
\caption{The accurate architecture begins with two convolutional feature
extractors. The extracted feature vectors are concatenated and compared by
a number of fully-connected layers. The inputs are two image patches and the
output is a single real number between 0 and 1, which we interpret as a measure
of similarity between the input images.}

\label{fig:architecture_accurate}
\end{center}
\end{figure}

The second architecture is derived from the first by replacing the cosine
similarity measure with a number of fully-connected layers (see
Figure~\ref{fig:architecture_accurate}). This architectural change increased
the running time, but decreased the error rate. The two sub-networks comprise a
number of convolutional layers, with a rectified linear unit following each
layer. The resulting two vectors are concatenated and forward-propagated
through a number of fully-connected layers followed by rectified linear units. The
last fully-connected layer produces a single number which, after being
transformed with the sigmoid nonlinearity, is interpreted as the similarity score
between the input patches.

We use the binary cross-entropy loss for training.  Let $s$ denote the output of
the network for one training example and $t$ denote the class of that
training example; $t=1$ if the example belongs to the positive class and $t=0$
if the example belongs to the negative class. The binary cross-entropy loss for
that example is defined as $t\log(s) + (1 - t)\log(1 - s)$.

The decision to use two different loss functions, one for each architecture,
was based on empirical evidence. While we would have preferred to use the same
loss function for both architectures, experiments showed that the binary
cross-entropy loss performed better than the hinge loss on the accurate
architecture. On the other hand, since the last step of the fast architecture
is the cosine similarity computation, a cross-entropy loss was not directly
applicable.

The hyperparameters of the accurate architecture are
the number of convolutional layers in each sub-network (\texttt{num\_conv\_layers}), 
the number of feature maps in each layer (\texttt{num\_conv\_feature\_maps}), 
the size of the convolution kernels (\texttt{conv\_kernel\_size}), 
the size of the input patch (\texttt{input\_patch\_size}), 
the number of units in each fully-connected layer (\texttt{num\_fc\_units}),
and the number of fully-connected layers (\texttt{num\_fc\_layers}). 

\subsection{Computing the Matching Cost}
\label{sec:matching_cost}

The output of the network is used to initialize the matching cost:
\begin{equation*}
C_{\text{CNN}}(\mathbf{p}, d) = -s(<\mathcal{P}^L(\mathbf{p}), \mathcal{P}^R(\mathbf{p}-\mathbf{d})>),
\end{equation*}
where $s(<\mathcal{P}^L(\mathbf{p}), \mathcal{P}^R(\mathbf{p}-\mathbf{d})>)$ is the
output of the network when run on input patches $\mathcal{P}^L(\mathbf{p})$ and
$\mathcal{P}^R(\mathbf{p}-\mathbf{d})$. The minus sign converts the similarity score to a
matching cost. 

To compute the entire matching cost tensor \(
C_{\text{CNN}}(\mathbf{p}, d) \) we would, naively, have to perform the forward
pass for each image location and each disparity under consideration. The
following three implementation details kept the running time manageable:
\begin{itemize}
\item The outputs of the two sub-networks need to be computed only once per
location, and do not need to be recomputed for every disparity under
consideration.

\item The output of the two sub-networks can be computed for all pixels in a
single forward pass by propagating full-resolution images, instead of small
image patches. Performing a single forward pass on the entire $w \times h$
image is faster than performing $w \cdot h$ forward passes on small patches
because many intermediate results can be reused.

\item The output of the fully-connected layers in the accurate architecture can
also be computed in a single forward pass. This is done by replacing each
fully-connected layer with a convolutional layer with \(1 \times 1\) kernels.
We still need to perform the forward pass for each disparity under
consideration; the maximum disparity $d$ is 228 for the KITTI data set and 400
for the Middlebury data set. As a result, the fully-connected part of the
network needs to be run \(d\) times, and is a bottleneck of the accurate
architecture.

\end{itemize}
To compute the matching cost of a pair of images, we run the sub-networks once
on each image and run the fully-connected layers $d$ times, where $d$ is the
maximum disparity under consideration. This insight was important in designing
the architecture of the network. We could have chosen an architecture where the
two images are concatenated before being presented to the network, but that
would imply a large cost at runtime because the whole network would need to be
run $d$ times. This insight also led to the development  of the fast
architecture, where the only layer that is run $d$ times is the dot product of
the feature vectors.

\section{Stereo Method}
\label{sec:stereo_method}

The raw outputs of the convolutional neural network are not enough to produce
accurate disparity maps, with errors particularly apparent in low-texture
regions and occluded areas. The quality of the disparity maps can be improved
by applying a series of post-processing steps referred to as the stereo method.
The stereo method we used was influenced by \citet{mei2011building} and
comprises cross-based cost aggregation, semiglobal matching, a left-right
consistency check, subpixel enhancement, a median, and a bilateral filter.

\subsection{Cross-based Cost Aggregation}

Information from neighboring pixels can be combined by averaging the matching
cost over a fixed window. This approach fails near depth discontinuities,
where the assumption of constant depth within a window is violated. We might
prefer a method that adaptively selects the neighborhood for each pixel, so
that support is collected only from pixels of the same physical object. In
cross-based cost aggregation \citep{zhang2009cross} we build a local
neighborhood around each location comprising pixels with similar image
intensity values with the hope that these pixels belong to the same object.

\begin{figure}[tb]
\begin{center}
\includegraphics[width=0.4\textwidth]{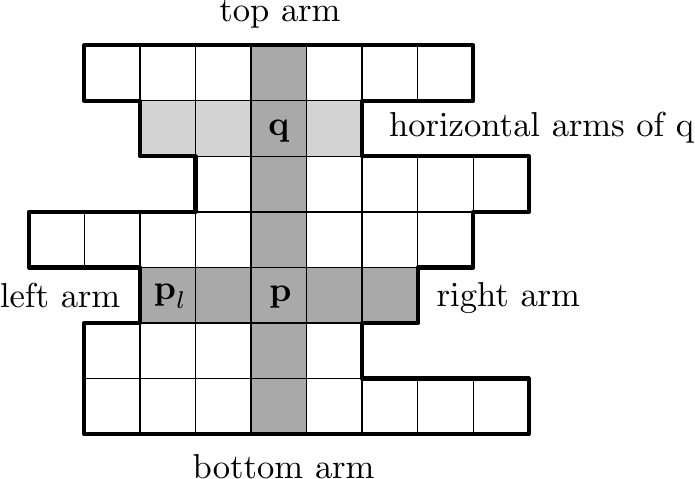}
\end{center}
\caption{The support region for position $\mathbf{p}$ is the union of
horizontal arms of all positions $\mathbf{q}$ on $\mathbf{p}$'s vertical arm.}
\label{fig:cross}
\end{figure}

The method begins by constructing an upright cross at each position; this cross
is used to define the local support region. The left arm $\mathbf{p}_l$ at
position $\mathbf{p}$ extends left as long as the following two conditions
hold: 

\begin{itemize}

\item $|I(\mathbf{p}) - I(\mathbf{p}_l)| < \texttt{cbca\_intensity}$; the image
intensities at positions $\mathbf{p}$ and $\mathbf{p}_l$ should be similar,
their difference should be less than \texttt{cbca\_intensity}.

\item $\|\mathbf{p} - \mathbf{p}_l\| < \texttt{cbca\_distance}$; the horizontal
distance (or vertical distance in case of top and bottom arms) between positions
$\mathbf{p}$ and $\mathbf{p}_l$ is less than \texttt{cbca\_distance} pixels.

\end{itemize}
The right, bottom, and top arms are constructed analogously. Once the four arms
are known, we can compute the support region $U(\mathbf{p})$ as the union of
horizontal arms of all positions $\mathbf{q}$ laying on $\mathbf{p}$'s
vertical arm (see Figure \ref{fig:cross}).

\citet{zhang2009cross} suggest that aggregation should consider the support
regions of both images in a stereo pair. Let $U^L$ and $U^R$ denote the support
regions in the left and right image. We define the combined support region
$U_d$ as
\begin{equation*}
U_d(\mathbf{p}) = \{\mathbf{q} | \mathbf{q} \in U^L(\mathbf{p}), \mathbf{q}-\mathbf{d}
\in U^R(\mathbf{p}-\mathbf{d})\}.
\end{equation*}

The matching cost is averaged over the combined support region:
\begin{align*}
C^0_{\text{CBCA}}(\mathbf{p}, d) &= C_{\text{CNN}}(\mathbf{p}, d), \\
C^i_{\text{CBCA}}(\mathbf{p}, d) &= \frac{1}{|U_d(\mathbf{p})|}
\sum_{\mathbf{q} \in U_d(\mathbf{p})} C^{i-1}_{\text{CBCA}}(\mathbf{q}, d),
\end{align*}
where $i$ is the iteration number. We repeat the averaging a number of times.
Since the support regions are overlapping, the results can change at each
iteration. We skip cross-based cost aggregation in the fast architecture
because it is not crucial for achieving a low error rate and because it is
relatively expensive to compute.

\subsection{Semiglobal Matching}

We refine the matching cost by enforcing smoothness constraints on the
disparity image. Following \citet{hirschmuller2008stereo}, we define an energy
function $E(D)$ that depends on the disparity image $D$:
\begin{multline*} 
E(D) = \sum_{\mathbf{p}} \biggl( C^4_{\text{CBCA}}(\mathbf{p}, D(\mathbf{p}))
+ \sum_{\mathbf{q} \in \mathcal{N}_{\mathbf{p}}} P_1 \cdot 1\{|D(\mathbf{p}) - D(\mathbf{q})| = 1\} \\
+ \sum_{\mathbf{q} \in \mathcal{N}_{\mathbf{p}}} P_2 \cdot 1\{|D(\mathbf{p}) - D(\mathbf{q})| > 1\} \biggr), 
\end{multline*}
where $1\{\cdot\}$ denotes the indicator function. The first term penalizes
disparities with high matching costs. The second term adds a penalty $P_1$ when
the disparity of neighboring pixels differ by one. The third term adds a
larger penalty $P_2$ when the neighboring disparities differ by more than one.

Rather than minimizing $E(D)$ in all directions simultaneously, we could
perform the minimization in a single direction with dynamic programming. This
solution would introduce unwanted streaking effects, since there would be no
incentive to make the disparity image smooth in the directions we are not
optimizing over.  In semiglobal matching we minimize the energy in a single
direction, repeat for several directions, and average to obtain the final
result.  Although \citet{hirschmuller2008stereo} suggested choosing sixteen
direction, we only optimized along the two horizontal and the two vertical
directions; adding the diagonal directions did not improve the accuracy of our
system.  To minimize $E(D)$ in direction $\mathbf{r}$, we define a matching
cost $C_{\mathbf{r}}(\mathbf{p}, d)$ with the following recurrence relation:
\begin{multline*} C_{\mathbf{r}}(\mathbf{p}, d) = C^4_{\text{CBCA}}(\mathbf{p},
d) - \min_k C_r(\mathbf{p} - \mathbf{r}, k) + \min\biggl\{ C_r(\mathbf{p} -
\mathbf{r}, d), C_r(\mathbf{p} - \mathbf{r}, d - 1) + P_1,\\ C_r(\mathbf{p} -
\mathbf{r}, d + 1) + P_1, \min_k C_{\mathbf{r}}(\mathbf{p} - \mathbf{r}, k) +
P_2 \biggr\}.  \end{multline*}
The second term is subtracted to prevent values of $C_\mathbf{r}(\mathbf{p},
d)$ from growing too large and does not affect the optimal disparity map. 

The penalty parameters $P_1$ and $P_2$ are set according to the image gradient
so that jumps in disparity coincide with edges in the image. Let $D_1 =
|I^L(\mathbf{p}) - I^L(\mathbf{p} - \mathbf{r})|$ and $D_2 = |I^R(\mathbf{p}-\mathbf{d})
- I^R(\mathbf{p}-\mathbf{d} - \mathbf{r})|$ be the difference in image intensity between
two neighboring positions in the direction we are optimizing over. We set
$P_1$ and $P_2$ according to the following rules:
\[ \begin{array}{lll} 
P_1 = \texttt{sgm\_P1}, &P_2 = \texttt{sgm\_P2} & 
\text{if $D_1 < \texttt{sgm\_D}, D_2 < \texttt{sgm\_D}$}; \\ 
P_1 = \texttt{sgm\_P1} / \texttt{sgm\_Q2}, &P_2 = \texttt{sgm\_P2} / \texttt{sgm\_Q2} & 
\text{if $D_1 \geq \texttt{sgm\_D}, D_2 \geq \texttt{sgm\_D}$}; \\ 
P_1 = \texttt{sgm\_P1} / \texttt{sgm\_Q1}, &P_2 = \texttt{sgm\_P2} / \texttt{sgm\_Q1} & 
\text{otherwise.} \\ 
\end{array} \]
The hyperparameters \texttt{sgm\_P1} and \texttt{sgm\_P2} set a base penalty
for discontinuities in the disparity map. The base penalty is reduced by a
factor of \texttt{sgm\_Q1} if one of $D_1$ or $D_2$ indicate a strong image
gradient or by a larger factor of \texttt{sgm\_Q2} if both $D_1$ and $D_2$
indicate a strong image gradient. The value of $P_1$ is further reduced by a
factor of \texttt{sgm\_V} when considering the two vertical directions; in the
ground truth, small changes in disparity are much more frequent in the vertical
directions than in the horizontal directions and should be penalised less. 

The final cost $C_\text{SGM}(\mathbf{p}, d)$ is computed by taking the
average across all four directions:

\begin{equation*} 
C_\text{SGM}(\mathbf{p}, d) = \frac{1}{4} \sum_{\mathbf{r}}
C_{\mathbf{r}}(\mathbf{p}, d). 
\end{equation*}

After semiglobal matching we repeat cross-based cost aggregation, as described
in the previous section. Hyperparameters \texttt{cbca\_num\_iterations\_1}
and \texttt{cbca\_num\_iterations\_2} determine the number of cross-based
cost aggregation iterations before and after semiglobal matching.

\subsection{Computing the Disparity Image}

The disparity image $D(\mathbf{p})$ is computed by the winner-takes-all
strategy, that is, by finding the disparity $d$ that minimizes $C(\mathbf{p},
d)$,
\begin{equation*}
D(\mathbf{p}) = \arg\!\min_d C(\mathbf{p}, d).
\end{equation*}

\subsubsection{Interpolation}

The interpolation steps attempt to resolve conflicts between the disparity
map predicted for the left image and the disparity map predicted for the right
image. Let $D^L$ denote the disparity map obtained by treating the left image
as the reference image---this was the case so far, that is, $D^L(\mathbf{p}) =
D(\mathbf{p})$---and let $D^R$ denote the disparity map obtained by treating
the right image as the reference image. $D^L$ and $D^R$ sometimes disagree on
what the correct disparity at a particular position should be. We detect these
conflicts by performing a left-right consistency check. We label each
position $\mathbf{p}$ by applying the following rules in turn:
\[ \begin{array}{ll} 
correct & \text{if $|d - D^R(\mathbf{p}-\mathbf{d})| \leq 1$ for $d = D^L(\mathbf{p})$}, \\ 
mismatch & \text{if $|d - D^R(\mathbf{p}-\mathbf{d})| \leq 1$ for any other $d$}, \\ 
occlusion & \text{otherwise}. \\ 
\end{array} \]
For positions marked as \emph{occlusion}, we want the new disparity value to
come from the background. We interpolate by moving left until we find a
position labeled \emph{correct} and use its value. For positions marked as
\emph{mismatch}, we find the nearest \emph{correct} pixels in 16 different
directions and use the median of their disparities for interpolation. 
We refer to the interpolated disparity map as $D_{\text{INT}}$.

\subsubsection{Subpixel Enhancement}

Subpixel enhancement provides an easy way to increase the resolution of a
stereo algorithm. We fit a quadratic curve through the neighboring costs to
obtain a new disparity image:
\begin{equation*}
D_{\text{SE}}(\mathbf{p}) = d - \frac {C_+ - C_-} {2 (C_+ - 2 C + C_-)},
\end{equation*}
where
$d = D_{\text{INT}}(\mathbf{p})$,
$C_- = C_{\text{SGM}}(\mathbf{p}, d - 1)$,
$C   = C_{\text{SGM}}(\mathbf{p}, d    )$, and
$C_+ = C_{\text{SGM}}(\mathbf{p}, d + 1)$.

\subsubsection{Refinement}

The final steps of the stereo method consist of a $5 \times 5$ median filter and the
following bilateral filter:
\begin{equation*} 
D_{\text{BF}}(\mathbf{p}) = \frac{1}{W(\mathbf{p})}
\sum_{\mathbf{q} \in \mathcal{N}_\mathbf{p}} D_{\text{SE}}(\mathbf{q}) \cdot
g(\|\mathbf{p} - \mathbf{q}\|) \cdot 1\{|I^L(\mathbf{p}) - I^L(\mathbf{q})|
< \texttt{blur\_threshold}\},
\end{equation*}
where $g(x)$ is the probability density function of a zero mean normal
distribution with standard deviation \texttt{blur\_sigma} and $W(\mathbf{p})$ is the
normalizing constant,
\begin{equation*}
W(\mathbf{p}) = \sum_{\mathbf{q} \in \mathcal{N}_\mathbf{p}} g(\|\mathbf{p} -
\mathbf{q}\|) \cdot 1\{|I^L(\mathbf{p}) - I^L(\mathbf{q})| <
\texttt{blur\_threshold}\}.
\end{equation*}
The role of the bilateral filter is to smooth the disparity map without
blurring the edges. $D_{\text{BF}}$ is the final output of our stereo method.

\section{Experiments}

We used three stereo data sets in our experiments: KITTI 2012, KITTI 2015, and
Middle\-bury.  The test set error rates reported in Tables
\ref{tbl:kitti2012_leaderboard}, \ref{tbl:kitti2015_leaderboard}, and
\ref{tbl:middlebury_leaderboard} were obtained by submitting the generated
disparity maps to the online evaluation servers.  All other error rates were
computed by splitting the data set in two, using one part for training and the
other for validation.

\subsection{KITTI Stereo Data Set}

The KITTI stereo data set \citep{Geiger2013IJRR,menze2015object} is a collection
of rectified image pairs taken from two video cameras mounted on the roof of
a car, roughly 54 centimeters apart. The images were recorded while driving in
and around the city of Karlsruhe, in sunny and cloudy weather, at daytime. The
images were taken at a resolution of $1240 \times 376$. A rotating laser
scanner mounted behind the left camera recorded ground truth depth, labeling
around 30\,\% of the image pixels. 

The ground truth disparities for the test set are withheld and an online
leaderboard is provided where researchers can evaluate their method on the test
set. Submissions are allowed once every three days. Error is measured as the
percentage of pixels where the true disparity and the predicted disparity
differ by more than three pixels. Translated into distance, this means that,
for example, the error tolerance is 3 centimeters for objects 2 meters from the
camera and 80 centimeters for objects 10 meters from the camera. 

\begin{table}[tb]
\begin{center}
\begin{tabular}{clllSS}\toprule
Rank & Method & & Setting & {Error} & {Runtime} \\\midrule
1 & \textbf{MC-CNN-acrt} & \textbf{Accurate architecture} & & 2.43 & 67 \\
2 & Displets & \citet{guney2015displets} & & 2.47 & 265 \\
3 & MC-CNN & \citet{Zbontar_2015_CVPR} & & 2.61 & 100 \\
4 & PRSM & \citet{vogel20153d} & F, MV & 2.78 & 300 \\
  & \textbf{MC-CNN-fst} & \textbf{Fast architecture} & & 2.82 & 0.8 \\
5 & SPS-StFl & \citet{yamaguchi2014efficient} & F, MS & 2.83 & 35 \\
6 & VC-SF & \citet{vogel2014view} & F, MV & 3.05 & 300 \\
7 & Deep Embed & \citet{deep_embed} & & 3.10 & 3 \\
8 & JSOSM & Unpublished work & & 3.15 & 105 \\
9 & OSF & \citet{menze2015object} & F & 3.28 & 3000 \\
10 & CoR & \citet{chakrabarti2014low} & & 3.30 & 6 \\\bottomrule
\end{tabular}
\caption{The highest ranking methods on the KITTI 2012 data set as of October
2015. The ``Setting'' column provides insight into how the disparity map is
computed: ``F'' indicates the use of optical flow, ``MV'' indicates more than
two temporally adjacent images, and ``MS'' indicates the use of epipolar
geometry for computing the optical flow. The ``Error'' column reports the
percentage of misclassified pixels and the ``Runtime'' column measures the
time, in seconds, required to process one pair of images. }

\label{tbl:kitti2012_leaderboard}
\end{center}
\end{table}

Two KITTI stereo data sets exist: KITTI 2012\footnote{The KITTI 2012
scoreboard:
\url{http://www.cvlibs.net/datasets/kitti/eval_stereo_flow.php?benchmark=stereo}}
and, the newer, KITTI 2015\footnote{The KITTI 2015 scoreboard:
\url{http://www.cvlibs.net/datasets/kitti/eval_scene_flow.php?benchmark=stereo}}.
For the task of computing stereo they are nearly identical, with the newer data
set improving some aspects of the optical flow task. The 2012 data set contains
194 training and 195 testing images, while the 2015 data set contains 200
training and 200 testing images. There is a subtle but important difference
introduced in the newer data set: vehicles in motion are densely labeled and
car glass is included in the evaluation. This emphasizes the method's
performance on reflective surfaces.

\begin{figure}[p]
\setlength\tabcolsep{2pt}
\begin{center}
\begin{tabular}{ccc}%
\rule{0pt}{4ex}Left input image &
Right input image &
Ground truth \\
\includegraphics[scale=0.5]{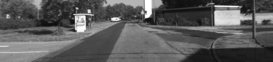} &
\includegraphics[scale=0.5]{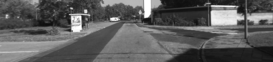} &
\includegraphics[scale=0.5]{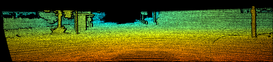}\\
\end{tabular}

\begin{tabular}{cc}
Census &
Error: 4.63\,\% \\
\includegraphics[scale=0.5]{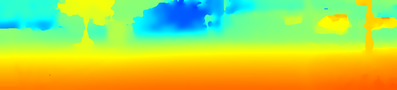} &
\includegraphics[scale=0.5]{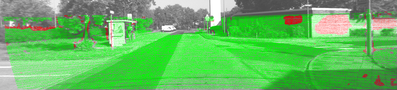}\\

Fast architecture &
Error: 1.01\,\% \\
\includegraphics[scale=0.5]{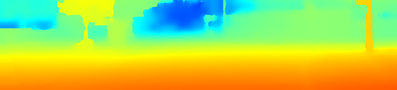} &
\includegraphics[scale=0.5]{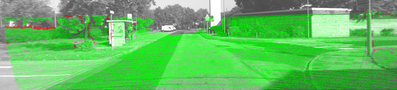} \\

Accurate architecture &
Error: 0.91\,\% \\
\includegraphics[scale=0.5]{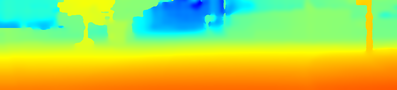} &
\includegraphics[scale=0.5]{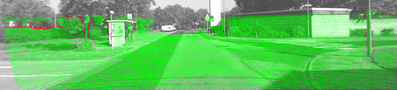}\\
\end{tabular}

\begin{tabular}{ccc}
\midrule
\rule{0pt}{4ex}Left input image &
Right input image &
Ground truth \\
\includegraphics[scale=0.5]{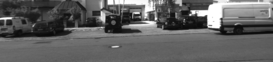} &
\includegraphics[scale=0.5]{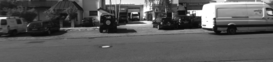} &
\includegraphics[scale=0.5]{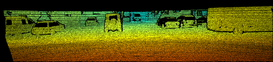}\\
\end{tabular}

\begin{tabular}{cc}
Census &
Error: 14.12\,\% \\
\includegraphics[scale=0.5]{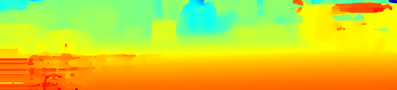} &
\includegraphics[scale=0.5]{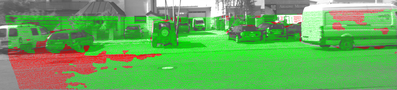}\\

Fast architecture &
Error: 1.54\,\% \\
\includegraphics[scale=0.5]{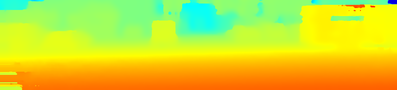} &
\includegraphics[scale=0.5]{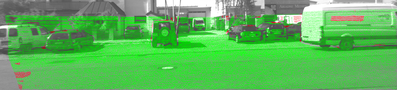} \\

Accurate architecture &
Error: 1.45\,\% \\
\includegraphics[scale=0.5]{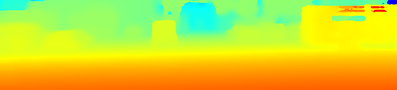} &
\includegraphics[scale=0.5]{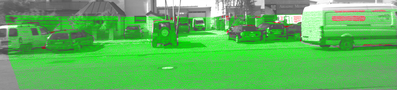}\\
\end{tabular}

\caption{Examples of predicted disparity maps on the KITTI 2012 data set. Note
how some regions of the image (the white wall in the top example, and the
asphalt in the bottom example) cause problems for the census transform. The
fast and the accurate architecture perform better, with the accurate
architecture making fewer mistakes on average.}

\label{fig:kitti2012}
\end{center}
\end{figure}

The best performing methods on the KITTI 2012 data set are listed in
Table~\ref{tbl:kitti2012_leaderboard}. Our accurate architecture ranks first
with an error rate of 2.43\,\%. Third place on the leaderboard is held by our
previous work \citep{Zbontar_2015_CVPR} with an error rate of 2.61\,\%. The two
changes that reduced the error from 2.61\,\% to 2.43\,\% were augmenting the
data set (see Section~\ref{sec:dataset_augmentation}) and doubling the number
of convolution layers while reducing the kernel size from $5 \times 5$ to $3
\times 3$. The method in second place \citep{guney2015displets} uses the
matching cost computed by our previous work \citep{Zbontar_2015_CVPR}. The
test error rate of the fast architecture is 2.82\,\%, which would be enough for
fifth place had the method been allowed to appear in the public leaderboard.
The running time for processing a single image pair is 67 seconds for the
accurate architecture and 0.8 seconds for the fast architecture.
Figure~\ref{fig:kitti2012} contains a pair of examples from the KITTI 2012 data
set, together with the predictions of our method.

\begin{table}[tb]
\setlength\tabcolsep{2pt}
\begin{center}
\begin{tabular}{cllcSS}\toprule
Rank & Method & & Setting & {Error} & {Runtime} \\\midrule
1 & \textbf{MC-CNN-acrt} & \textbf{Accurate architecture} & & 3.89 & 67\\
  & \textbf{MC-CNN-fst} & \textbf{Fast architecture} & & 4.62 & 0.8\\
2 & SPS-St & \citet{yamaguchi2014efficient} & & 5.31 & 2\\
3 & OSF & \citet{menze2015object} & F & 5.79 & 3000\\
4 & PR-Sceneflow & \citet{vogel2013piecewise} & F & 6.24 & 150\\
5 & SGM+C+NL & \citet{hirschmuller2008stereo,sun2014quantitative} & F & 6.84 & 270\\
6 & SGM+LDOF & \citet{hirschmuller2008stereo,brox2011large} & F & 6.84 & 86\\
7 & SGM+SF & \citet{hirschmuller2008stereo,hornacek2014sphereflow} & F & 6.84 & 2700\\
8 & ELAS & \citet{geiger2011efficient} & & 9.72 & 0.3\\
9 & OCV-SGBM & \citet{hirschmuller2008stereo} & & 10.86 & 1.1\\
10 & SDM & \citet{kostkova2003stratified} & & 11.96 & 60\\\bottomrule
\end{tabular}
\caption{The leading submission on the KITTI 2015 leaderboard as of October 2015. 
The ``Setting'', ``Error'', and ``Runtime'' columns have the same
meaning as in Table~\ref{tbl:kitti2012_leaderboard}.}
\label{tbl:kitti2015_leaderboard}
\end{center}
\end{table}

\begin{figure}[p]
\setlength\tabcolsep{2pt}
\begin{center}
\begin{tabular}{ccc}%
\rule{0pt}{4ex}Left input image &
Right input image &
Ground truth \\
\includegraphics[scale=0.5]{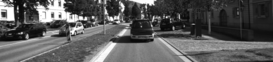} &
\includegraphics[scale=0.5]{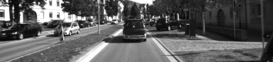} &
\includegraphics[scale=0.5]{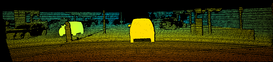}
\end{tabular}

\begin{tabular}{cc}
Census &
Error: 4.58\,\% \\
\includegraphics[scale=0.5]{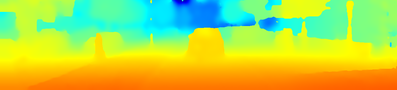} &
\includegraphics[scale=0.5]{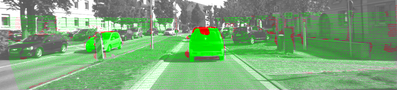}\\

Fast architecture &
Error: 2.79\,\% \\
\includegraphics[scale=0.5]{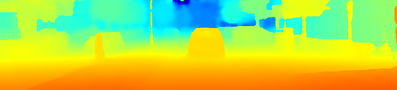} &
\includegraphics[scale=0.5]{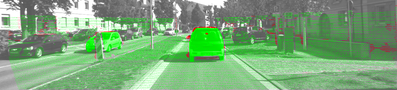} \\

Accurate architecture &
Error: 2.36\,\% \\
\includegraphics[scale=0.5]{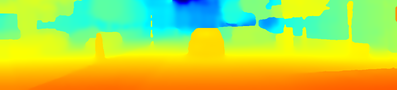} &
\includegraphics[scale=0.5]{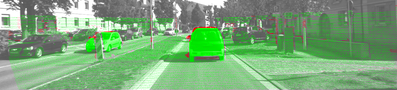}\\
\end{tabular}

\begin{tabular}{ccc}%
\midrule
\rule{0pt}{4ex}Left input image &
Right input image &
Ground truth \\
\includegraphics[scale=0.5]{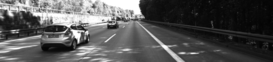} &
\includegraphics[scale=0.5]{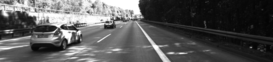} &
\includegraphics[scale=0.5]{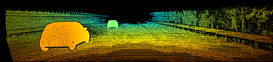}\\
\end{tabular}

\begin{tabular}{cc}
Census &
Error: 5.25\,\% \\
\includegraphics[scale=0.5]{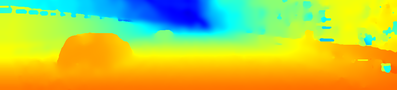} &
\includegraphics[scale=0.5]{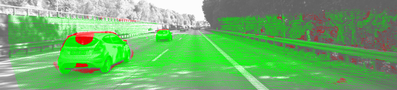}\\

Fast architecture &
Error: 3.91\,\% \\
\includegraphics[scale=0.5]{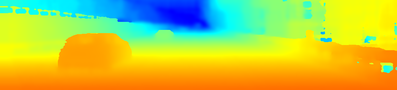} &
\includegraphics[scale=0.5]{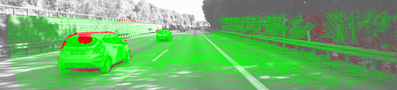} \\

Accurate architecture &
Error: 3.73\,\% \\
\includegraphics[scale=0.5]{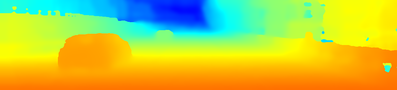} &
\includegraphics[scale=0.5]{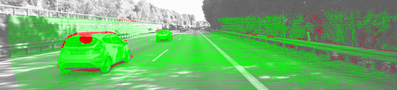}\\
\end{tabular}

\caption{Examples of predictions on the KITTI 2015 data set. Observe that
vehicles in motion are labeled densely in the KITTI 2015 data set. }

\label{fig:kitti2015}
\end{center}
\end{figure}

Table~\ref{tbl:kitti2015_leaderboard} presents the frontrunners on the KITTI
2015 data sets. The error rates of our methods are 3.89\,\% for the accurate
architecture and 4.46\,\% for the fast architecture, occupying first and second
place on the leaderboard. Since one submission per paper is allowed, only the
result of the accurate architecture appears on the public leaderboard. See
Figure~\ref{fig:kitti2015} for the disparity maps produced by our method on the
KITTI 2015 data set.

\subsection{Middlebury Stereo Data Set}

The image pairs of the Middlebury stereo data set are indoor scenes taken under
controlled lighting conditions. Structured light was used to measure the true
disparities with higher density and precision than in the KITTI data set. The
data sets were published in five separate works in the years 2001, 2003, 2005,
2006, and 2014 \citep{scharstein2002taxonomy, scharstein2003high,
scharstein2007learning, hirschmuller2007evaluation, scharstein2014high}. In
this paper, we refer to the Middlebury data set as the concatenation of all five
data sets; a summary of each is presented in Table~\ref{tbl:middlebury}.

\begin{table}[tb] 
\begin{center} 
\begin{tabular}{cccc}
\toprule 
Year & Number of Image Pairs & Resolution & Maximum Disparity \\\midrule 
2001 &  8 & \(380 \times 430 \) & 30 \\ 
2003 &  2 & \(1800 \times 1500 \) & 220 \\ 
2005 &  6 & \(1400 \times 1100 \) & 230 \\ 
2006 & 21 & \(1400 \times 1100 \) & 230 \\ 
2014 & 23 & \(3000 \times 2000 \) & 800 \\\bottomrule 
\end{tabular} 
\caption{A summary of the five Middlebury stereo data sets. The column ``Number
of Image Pairs'' counts only the image pairs for which ground truth is
available. The 2005 and 2014 data sets additionally contain a number of image
pairs with ground truth disparities withheld; these image pairs constitute the test
set.}

\label{tbl:middlebury}
\end{center} 
\end{table}

Each scene in the 2005, 2006, and 2014 data sets was taken under a number of
lighting conditions and shutter exposures, with a typical image pair taken
under four lighting conditions and seven exposure settings for a total of 28
images of the same scene. 

An online leaderboard\footnote{The Middlebury scoreboard:
\url{http://vision.middlebury.edu/stereo/eval3/}}, similar to the one provided
by KITTI, displays a ranked list of all submitted methods. Participants have
only one opportunity to submit their results on the test set to the public
leaderboard. This rule is stricter than the one on the KITTI data set, where
submissions are allowed every three days. The test set contains 15 images
borrowed from the 2005 and 2014 data sets. 

The data set is provided in full, half, and quarter resolution. The error is
computed at full resolution; if the method outputs half or quarter resolution
disparity maps, they are upsampled before the error is computed. We chose to
run our method on half resolution images because of the limited size of the
graphic card's memory available.

Rectifying a pair of images using standard calibration procedures, like the
ones present in the OpenCV library, results in vertical disparity errors of up
to nine pixels on the Middlebury data set \citep{scharstein2014high}. Each
stereo pair in the 2014 data set is rectified twice: once using a standard,
imperfect approach, and once using precise 2D correspondences for perfect
rectification \citep{scharstein2014high}. We train the network on imperfectly
rectified image pairs, since only two of the fifteen test images
(\emph{Australia} and \emph{Crusade}) are rectified perfectly. 

The error is measured as the percentage of pixels where the true disparity and
the predicted disparity differ by more than two pixels; this corresponds to an
error tolerance of one pixel at half resolution. The error on the evaluation
server is, by default, computed only on non-occluded pixels. The final error
reported online is the weighted average over the fifteen test images, with
weights set by the authors of the data set.

\begin{table}[tb]
\begin{center}
\begin{tabular}{clllSS}\toprule
Rank & Method & & Resolution & {Error} & {Runtime} \\\midrule
1 & \textbf{MC-CNN-acrt} & \textbf{Accurate architecture} & Half & 8.29 & 150 \\
2 & MeshStereo & \citet{zhang2015meshstereo} & Half & 13.4 & 65.3 \\
3 & LCU & Unpublished work & Quarter & 17.0 & 6567 \\
4 & TMAP & \citet{psota2015map} & Half & 17.1 & 2435 \\
5 & IDR & \citet{kowalczuk2013real} & Half & 18.4 & 0.49 \\
6 & SGM & \citet{hirschmuller2008stereo} & Half & 18.7 & 9.90 \\
7 & LPS & \citet{sinha2014efficient} & Half & 19.4 & 9.52 \\
8 & LPS & \citet{sinha2014efficient} & Full & 20.3 & 25.8 \\
9 & SGM & \citet{hirschmuller2008stereo} & Quarter & 21.2 & 1.48 \\
10 & SNCC & \citet{einecke2010two} & Half & 22.2 & 1.38 \\\bottomrule
\end{tabular}
\caption{The top ten methods on the Middlebury stereo data set as of October
2015. The ``Error'' column is the weighted average error after upsampling to
full resolution and ``Runtime'' is the time, in seconds, required to process
one pair of images.}

\label{tbl:middlebury_leaderboard}
\end{center}
\end{table}

\begin{figure}[p]
\setlength\tabcolsep{2pt}
\begin{center}
\begin{tabular}{ccc}%
\rule{0pt}{4ex}Left input image &
Right input image &
Ground truth \\
\includegraphics[scale=0.5]{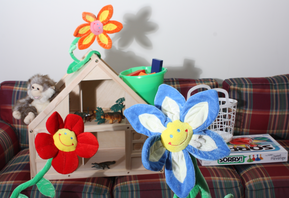} &
\includegraphics[scale=0.5]{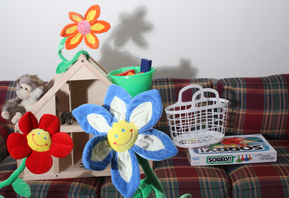} &
\includegraphics[scale=0.5]{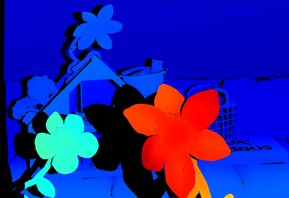}\\
\end{tabular}

\begin{tabular}{cc}
Census &
Error: 34.65\,\% \\
\includegraphics[scale=0.5]{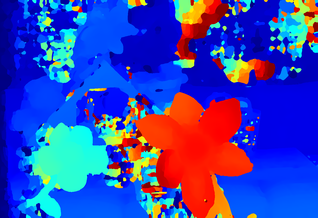} &
\includegraphics[scale=0.5]{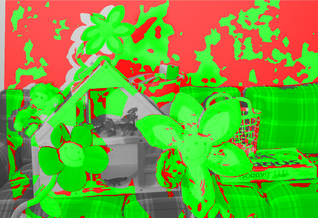}\\

Fast architecture &
Error: 16.19\,\% \\
\includegraphics[scale=0.5]{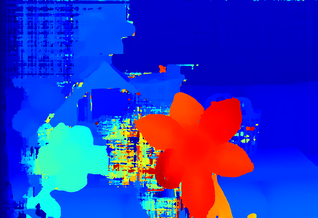} &
\includegraphics[scale=0.5]{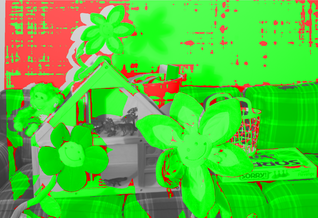} \\

Accurate architecture &
Error: 7.25\,\% \\
\includegraphics[scale=0.5]{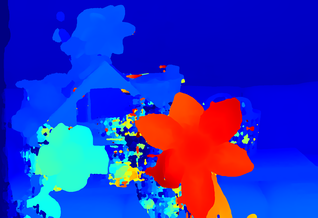} &
\includegraphics[scale=0.5]{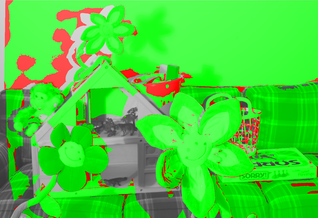}\\
\end{tabular}

\caption{An example of a particularly difficult image pair from the Middlebury
data set; the white wall in the background is practically textureless. The
accurate architecture is able to classify most of it correctly. The fast
architecture doesn't do as well but still performs better than census.}

\label{fig:middlebury}
\end{center}
\end{figure}

Table~\ref{tbl:middlebury_leaderboard} contains a snapshot of the third, and
newest, version of the Middlebury leaderboard. Our method ranks first with an
error rate of 8.29\,\% and a substantial lead over the second placed MeshStereo
method, whose error rate is 13.4\,\%. See Figure~\ref{fig:middlebury} for
disparity maps produced by our method on one image pair from the Middlebury
data set.

\subsection{Details of Learning}

\begin{table}[tb]
\begin{center}
\begin{tabular}{l ccc ccc}\toprule

& \multicolumn{2}{c}{KITTI 2012} &
\multicolumn{2}{c}{KITTI 2015} &
\multicolumn{2}{c}{Middlebury} \\
\cmidrule(lr){2-3}
\cmidrule(lr){4-5}
\cmidrule(lr){6-7}
Hyperparameter & fst & acrt & fst & acrt & fst & acrt \\\midrule
\texttt{input\_patch\_size} & \( 9 \times 9 \) & \( 9 \times 9 \) & \( 9 \times 9 \) & \( 9 \times 9 \) & \( 11 \times 11 \) & \( 11 \times 11 \) \\
\texttt{num\_conv\_layers} & 4 & 4 & 4 & 4 & 5 & 5 \\
\texttt{num\_conv\_feature\_maps} & 64 & 112 & 64 & 112 & 64 & 112\\
\texttt{conv\_kernel\_size} & 3 & 3 & 3 & 3 & 3 & 3 \\
\texttt{num\_fc\_layers} & & 4 & & 4 & & 3 \\
\texttt{num\_fc\_units} & & 384 & & 384 & & 384 \\
\texttt{dataset\_neg\_low} & 4 & 4 & 4 & 4 & 1.5 & 1.5 \\
\texttt{dataset\_neg\_high} & 10 & 10 & 10 & 10 & 6 & 18 \\
\texttt{dataset\_pos} & 1 & 1 & 1 & 1 & 0.5 & 0.5 \\
\texttt{cbca\_intensity} & & 0.13 & & 0.03 & & 0.02 \\
\texttt{cbca\_distance} & & 5 & & 5 & & 14 \\
\texttt{cbca\_num\_iterations\_1} & & 2 & & 2 & & 2 \\
\texttt{cbca\_num\_iterations\_2} & & 0 & & 4 & & 16 \\
\texttt{sgm\_P1} & 4 & 1.32 & 2.3 & 2.3 & 2.3 & 1.3 \\
\texttt{sgm\_P2} & 223 & 32 & 42.3 & 55.8 & 55.9 & 18.1 \\
\texttt{sgm\_Q1} & 3 & 3 & 3 & 3 & 4 & 4.5 \\
\texttt{sgm\_Q2} & 7.5 & 6 & 6 & 6 & 8 & 9 \\
\texttt{sgm\_V} & 1.5 & 2 & 1.25 & 1.75 & 1.5 & 2.75 \\
\texttt{sgm\_D} & 0.02 & 0.08 & 0.08 & 0.08 & 0.08 & 0.13 \\
\texttt{blur\_sigma} & 7.74 & 6 & 4.64 & 6 & 6 & 1.7 \\
\texttt{blur\_threshold} & 5 & 6 & 5 & 5 & 2 & 2 \\\bottomrule

\end{tabular}
\caption{The hyperparameter values we used for the fast and accurate
architectures (abbreviated ``fst'' and ``acrt'').  Note that hyperparameters
concerning image intensity values (\texttt{cbca\_intensity} and
\texttt{sgm\_D}) apply to the preprocessed images and not to raw images with
intensity values in the range from 0 to 255. }

\label{tbl:hyperparameters_default}
\end{center}
\end{table}

We construct a binary classification data set from all available image pairs
in the training set. The data set contains 25 million examples on the KITTI
2012, 17 million examples on the KITTI 2015, and 38 million examples on the
Middlebury data set. 

At training time, the input to the network was a batch of 128 pairs of image
patches. At test time, the input was the entire left and right image. We could
have used entire images during training as well, as it would allow us to
implement the speed optimizations described in Section~\ref{sec:matching_cost}.
There were several reasons why we preferred to train on image patches: it was
easier to control the batch size, the examples could be shuffled so that one
batch contained patches from several different images, and it was easier to
maintain the same number of positive and negative examples within a batch. 

We minimized the loss using mini-batch gradient descent with the momentum term
set to 0.9. We trained for 14 epochs with the learning rate initially set to
0.003 for the accurate architecture and 0.002 for the fast architecture. The
learning rate was decreased by a factor of 10 on the 11$^{\text{th}}$ epoch.
The number of epochs, the initial learning rate, and the learning rate decrease
schedule where treated as hyperparameters and were optimized with
cross-validation. Each image was preprocessed by subtracting the mean and
dividing by the standard deviation of its pixel intensity values. The left and
right image of a stereo pair were preprocessed separately. Our initial
experiments suggested that using color information does not improve the quality
of the disparity maps; therefore, we converted all color images to grayscale.

The post-processing steps of the stereo method were implemented in CUDA
\citep{nickolls2008scalable}, the network training was done with the Torch
environment \citep{collobert2011torch7} using the convolution routines from
the cuDNN library \citep{chetlur2014cudnn}. The OpenCV library
\citep{opencv_library} was used for the affine transformation in the data
augmentation step. 

The hyperparameters where optimized with manual search and simple scripts that
helped automate the process. The hyperparameters we selected are shown in
Table~\ref{tbl:hyperparameters_default}.

\subsection{Data Set Augmentation}
\label{sec:dataset_augmentation}

Augmenting the data set by repeatedly transforming the training examples is a
commonly employed technique to reduce the network's generalization error. The
transformations are applied at training time and do not affect the runtime
performance. We randomly rotate, scale and shear the training patches; we also
change their brightness and contrast.  Since the transformations are applied to
patches after they have been extracted from the images, the data
augmentation step does not alter the ground truth disparity map or ruin the
rectification.

The parameters of the transformation are chosen randomly for each pair of
patches, and after one epoch of training, when the same example is being
presented to the network for the second time, new random parameters are
selected.  We choose slightly different transformation parameters for the left
and right image; for example, we would rotate the left patch by 10 degrees and
the right by 14. Different data sets benefited from different types of
transformations and, in some cases, using the wrong transformations increased
the error. 

On the Middlebury data set we took advantage of the fact that the
images were taken under different lighting conditions and different shutter
exposures by training on all available images.  
The same data set augmentation parameters were used for the KITTI 2012 and
KITTI 2015 data sets.

The Middlebury test data sets contains two images worth mentioning:
\emph{Classroom}, where the right image is underexposed and, therefore, darker
than the left; and \emph{Djembe}, where the left and right images were taken
under different light conditions. To handle these two cases we train, 20\,\% of
the time, on images where either the shutter exposure or the arrangements of
lights are different for the left and right image. 

We combat imperfect rectification on the Middlebury data set by including a
small vertical disparity between the left and right image patches. 

\begin{table}[tb]
\begin{center}
\begin{tabular}{l cc cc}\toprule

& \multicolumn{2}{c}{KITTI 2012} &
\multicolumn{2}{c}{Middlebury}\\
\cmidrule(lr){2-3}
\cmidrule(lr){4-5}
Hyperparameter & Range & Error & Range & Error \\\midrule

\texttt{rotate} & \([-7, 7] \) & 2.65 & \([-28, 28]\) & 7.99 \\
\texttt{scale} & & & \([0.8, 1]\) & 8.17 \\
\texttt{horizontal\_scale} & \([0.9, 1]\) & 2.62 & \([0.8, 1]\) & 8.08 \\
\texttt{horizontal\_shear} & \([0, 0.1]\) & 2.61 & \([0, 0.1]\) & 7.91 \\
\texttt{brightness} & \([0, 0.7]\) & 2.61 & \([0, 1.3]\) & 8.16 \\
\texttt{contrast} & \([1, 1.3]\) & 2.63 & \([1, 1.1]\) & 7.95 \\
\texttt{vertical\_disparity} & & & \([0, 1]\) & 8.05 \\
\texttt{rotate\_diff} & & & \([-3, 3]\) & 8.00 \\
\texttt{horizontal\_scale\_diff} & & & \([0.9, 1]\) & 7.97\\
\texttt{horizontal\_shear\_diff} & & & \([0, 0.3]\) & 8.05\\
\texttt{brightness\_diff} & \([0, 0.3] \) & 2.63 & \([0, 0.7]\) & 7.92 \\
\texttt{contrast\_diff} & & & \([1, 1.1]\) & 8.01 \\\midrule
No data set augmentation & & 2.73 & & 8.75 \\
Full data set augmentation & & 2.61 & & 7.91 \\\bottomrule
\end{tabular}
\caption{The hyperparameters governing data augmentation and how they affect
the validation error. The ``Error'' column reports the validation error when a
particular data augmentation step is not used. The last two rows report
validation errors with and without data augmentation. For example, the
validation error on the KITTI 2012 is 2.73\,\% if no data augmentation is used,
2.65\,\% if all steps except rotation are used, and 2.61\,\% if all data
augmentation steps are used. }

\label{tbl:da_params}
\end{center}
\end{table}

Before describing the steps of data augmentation, let us introduce some
notation: in the following, a word in \texttt{typewriter} is used to denote the
name of a hyperparameter defining a set, while the same word in \textit{italic}
is used to denote a number drawn randomly from that set.  For example,
\texttt{rotate} is a hyperparameter defining the set of possible rotations and
\textit{rotate} is a number drawn randomly from that set. 
The steps of data augmentation are presented in the following list:
\begin{itemize}
\item Rotate the left patch by \textit{rotate} degrees and the right patch by
\(\textit{rotate} + \textit{rotate\_diff}\) degrees.

\item Scale the left patch by \textit{scale} and the right patch by 
\(\textit{scale} \cdot \textit{scale\_diff} \).
\item Scale the left patch in the horizontal direction by \textit{horizontal\_scale} and
the right patch by \(\textit{horizontal\_scale} \cdot \textit{horizontal\_scale\_diff}\).
\item Shear the left patch in the horizontal direction by \textit{horizontal\_shear} and
the right patch by \(\textit{horizontal\_shear} + \textit{horizontal\_shear\_diff}\).
\item Translate the right patch in the vertical direction by \textit{vertical\_disparity}.
\item Adjust the brightness and contrast by setting the left and right image patches to:
\begin{align*}
\mathcal{P}^L &\leftarrow \mathcal{P}^L \cdot \textit{contrast} + \textit{brightness} \text{ and} \\
\mathcal{P}^R &\leftarrow \mathcal{P}^R \cdot (\textit{contrast} \cdot \textit{contrast\_diff}) + 
(\textit{brightness} + \textit{brightness\_diff}),
\end{align*}
with addition and multiplication carried out element-wise where appropriate.

\end{itemize}
Table~\ref{tbl:da_params} contains the hyperparameters used and measures how
each data augmentation step affected the validation error. 

Data augmentation reduced the validation error from 2.73\,\% to 2.61\,\% on the
KITTI 2012 data set and from 8.75\,\% to 7.91\,\% on the Middlebury data set.

\subsection{Runtime}

We measure the runtime of our implementation on a computer with a NVIDIA Titan
X graphics processor unit. Table~\ref{tbl:runtime} contains the runtime
measurements across a range of hyperparameter settings for three data sets:
KITTI, Middlebury half resolution, and a new, fictitious data set, called
Tiny, which we use to demonstrate the performance of our method on the kind of
images typically used for autonomous driving or robotics. The sizes
of images we measured the runtime on were: 1242 \(\times\) 350 with 228
disparity levels for the KITTI data set, 1500 \(\times\) 1000 with 200 disparity
levels for the Middlebury data set, and 320 \(\times\) 240 with 32 disparity
levels for the Tiny data set. 

\begin{table}[p]
\begin{center}
\begin{tabular}{lc cc cc cc}
\toprule
& & \multicolumn{2}{c}{KITTI} &
\multicolumn{2}{c}{Middlebury} &
\multicolumn{2}{c}{Tiny} \\
\cmidrule(lr){3-4}
\cmidrule(lr){5-6}
\cmidrule(lr){7-8}
Hyperparameter & & fst & acrt & fst & acrt & fst & acrt \\\midrule
\multirow{6}{*}{\texttt{num\_conv\_layers}}
& 1 & 0.23 & 66.1 & 0.70 & 74.9 & 0.01 & 1.7 \\
& 2 & 0.26 & 66.2 & 0.82 & 75.2 & 0.01 & 1.8 \\
& 3 & 0.30 & 66.3 & 0.97 & 75.4 & 0.02 & 1.8 \\
& 4 & 0.34 & 66.4 & 1.11 & 75.6 & 0.03 & 1.8 \\
& 5 & 0.38 & 66.5 & 1.24 & 75.7 & 0.03 & 1.9 \\
& 6 & 0.42 & 66.7 & 1.37 & 76.0 & 0.04 & 1.9 \\\midrule
\multirow{8}{*}{\texttt{num\_conv\_feature\_maps}}
& 16 & 0.09 & 59.4 & 0.27 & 64.8 & 0.01 & 1.6 \\
& 32 & 0.15 & 60.4 & 0.51 & 66.2 & 0.01 & 1.6 \\
& 48 & 0.25 & 61.5 & 0.94 & 68.2 & 0.02 & 1.7 \\
& 64 & 0.34 & 62.7 & 1.24 & 70.0 & 0.03 & 1.7 \\
& 80 & 0.44 & 64.0 & 1.63 & 72.0 & 0.04 & 1.8 \\
& 96 & 0.53 & 65.3 & 1.93 & 73.9 & 0.04 & 1.8 \\
& 112 & 0.61 & 66.4 & 2.28 & 75.7 & 0.05 & 1.8 \\
& 128 & 0.71 & 67.7 & 2.61 & 77.8 & 0.06 & 1.9 \\\midrule
\multirow{5}{*}{\texttt{num\_fc\_layers}}
& 1 & & 16.3 & & 25.3 & & 0.5 \\
& 2 & & 32.9 & & 50.7 & & 0.9 \\
& 3 & & 49.6 & & 75.7 & & 1.4 \\
& 4 & & 66.4 & & 101.2 & & 1.8 \\
& 5 & & 82.9 & & 126.4 & & 2.3 \\\midrule
\multirow{4}{*}{\texttt{num\_fc\_units}}
& 128 & & 17.4 & & 21.4 & & 0.6 \\
& 256 & & 38.5 & & 44.9 & & 1.1 \\
& 384 & & 66.4 & & 75.7 & & 1.8 \\
& 512 & & 101.0 & & 113.3 & & 2.7 \\\midrule
No stereo method & & 0.34 & 66.4 & 1.24 & 75.7 & 0.03 & 1.8 \\
Full stereo method & & 0.78 & 67.1 & 2.03 & 84.8 & 0.06 & 1.9 \\
\bottomrule
\end{tabular}
\caption{The time, in seconds, required to compute the matching cost, that is,
the time spent in the convolutional neural network without any post-processing
steps.  The time does include computing the matching cost twice: once when the
left image is taken to be the reference image and once when the right image is
taken to be the reference image.  We measure the runtime as a function of four
hyperparameters controlling the network architecture; for example, the first
six rows contain the runtime as the number of convolutional layers in the
network increases from one to six.  The last row of the table contains the
running time for the entire method, including the post-processing steps. As
before, we abbreviate the fast and accurate architectures as ``fst'' and
``acrt''.}

\label{tbl:runtime}
\end{center}
\end{table}

Table~\ref{tbl:runtime} reveals that the fast
architecture is up to 90 times faster than the accurate architecture.
Furthermore, the running times of the fast architecture are 0.78 seconds on
KITTI, 2.03 seconds on Middlebury, and 0.06 seconds on the Tiny data set.  We
can also see that the fully-connected layers are responsible for most of the
runtime in the accurate architecture, as the hyperparameters controlling the
number of convolutional layer and the number of feature maps have only a
small effect on the runtime.

Training times depended on the size of the data set and the architecture, but
never exceeded two days.

\subsection{Matching Cost}

We argue that the low error rate of our method is due to the convolutional
neural network and not a superior stereo method. We verify this claim by
replacing the convolutional neural network with three standard approaches for
computing the matching cost:

\begin{itemize}

\item \emph{The sum of absolute differences} computes the matching
cost according to Equation~\eqref{eqn:C_sad}, that is, the matching cost
between two image patches is computed by summing the absolute differences in
image intensities between corresponding locations. We used \(9 \times 9\)
patches.

\item \emph{The census transform}~\citep{zabih1994non} represents each image
position as a bit vector. The size of this vector is a hyperparameter whose
value, after examining several, we set to 81. The vector is computed
by cropping a \(9 \times 9 \) image patch centered around the position of
interest and comparing the intensity values of each pixel in the patch to the
intensity value of the pixel in the center. When the center pixel is brighter
the corresponding bit is set. The matching cost is computed as the hamming
distance between two census transformed vectors. 

\item \emph{Normalized cross-correlation} is a window-based method defined
with the following equation:
\begin{equation*}
C_{NCC}(\mathbf{p}, d) = \frac{\sum_{\mathbf{q} \in \mathcal{N}_{\mathbf{p}}} I^L(\mathbf{q}) I^R(\mathbf{q} - \mathbf{d})}
{\sqrt{\sum_{\mathbf{q} \in \mathcal{N}_{\mathbf{p}}} I^L(\mathbf{q})^2 \sum_{\mathbf{q} \in \mathcal{N}_{\mathbf{p}}} I^R(\mathbf{q} - \mathbf{d})^2 }}.
\end{equation*}
The normalized cross-correlation matching cost computes the cosine similarity
between the left and right image patch, when the left and right image patches
are viewed as vectors instead of matrices. This is the same function that
is computed in the last two layers of the fast architecture (normalization and
dot product).  The neighbourhood $\mathcal{N}_{\mathbf{p}}$ was set to a square
$11 \times 11$ window around $\mathbf{p}$.


\end{itemize}
The ``sad'', ``cens'', and ``ncc'' columns of Table~\ref{tbl:sm_skip} contain
the results of the sum of absolute differences, the census transform, and
normalized cross-correlation on the KITTI 2012, KITTI 2015, and Middlebury data
sets. The validation errors in the last rows of Table~\ref{tbl:sm_skip} should
be used to compare the five methods. On all three data sets the accurate
architecture performs best, followed by the fast architecture, which in turn is
followed by the census transform. These are the three best performing methods
on all three data sets. Their error rates are 2.61\,\%, 3.02\,\%, and 4.90\,\%
on KITTI 2012; 3.25\,\%, 3.99\,\%, and 5.03\,\% on KITTI 2015; and 7.91\,\%,
9.87\,\%, and 16.72\,\% on Middlebury. The sum of absolute differences and the
normalized cross-correlation matching costs produce disparity maps with larger
errors. For a visual comparison of our method and the census transform see
Figures~\ref{fig:kitti2012}, \ref{fig:kitti2015}, and \ref{fig:middlebury}.

\begin{table}[p]
\begin{center}
\begin{tabular}{lccccc}\toprule
& \multicolumn{5}{c}{KITTI 2012} \\
\cmidrule(lr){2-6}
& fst & acrt & sad & cens & ncc \\\midrule
Cross-based cost aggregation & 3.02 & 2.73 & 8.22 & 5.21 & 8.93 \\
Semiglobal matching          & 8.78 & 4.26 & 19.58 & 8.84 & 10.72 \\
Interpolation                & 3.48 & 2.96 & 9.21 & 5.96 & 11.16 \\
Subpixel Enhancement         & 3.03 & 2.65 & 8.16 & 4.95 & 8.93 \\
Median filter                & 3.03 & 2.63 & 8.16 & 4.92 & 9.00 \\
Bilateral filter             & 3.26 & 2.79 & 8.75 & 5.70 & 9.76 \\\midrule
No stereo method             & 15.70 & 13.49 & 32.30 & 53.55 & 22.21 \\
Full stereo method           & 3.02 & 2.61 & 8.16 & 4.90 & 8.93 \\
\end{tabular}
\begin{tabular}{l ccccc}\toprule
& \multicolumn{5}{c}{KITTI 2015} \\
\cmidrule(lr){2-6}
& fst & acrt & sad & cens & ncc \\\midrule
Cross-based cost aggregation & 3.99 & 3.39 & 9.94 & 5.20 & 8.89 \\
Semiglobal matching          & 8.40 & 4.51 & 19.80 & 7.25 & 9.36 \\
Interpolation                & 4.47 & 3.33 & 10.39 & 5.83 & 10.98 \\
Subpixel Enhancement         & 4.02 & 3.28 & 9.44 & 5.03 & 8.91 \\
Median filter                & 4.05 & 3.25 & 9.44 & 5.05 & 8.96 \\
Bilateral filter             & 4.20 & 3.43 & 9.95 & 5.84 & 9.77 \\\midrule
No stereo method             & 15.66 & 13.38 & 30.67 & 50.35 & 18.95 \\
Full stereo method           & 3.99 & 3.25 & 9.44 & 5.03 & 8.89 \\
\end{tabular}
\begin{tabular}{l ccccc}\toprule
& \multicolumn{5}{c}{Middlebury} \\
\cmidrule(lr){2-6}
& fst & acrt & sad & cens & ncc \\\midrule
Cross-based cost aggregation & 9.87  & 10.63 & 43.09 & 29.28 & 33.89 \\
Semiglobal matching          & 25.50 & 11.99 & 51.25 & 19.51 & 35.36 \\
Interpolation                & 9.87  & 7.91  & 41.86 & 16.72 & 33.89 \\
Subpixel Enhancement         & 10.29 & 8.44  & 42.71 & 17.18 & 34.12 \\
Median filter                & 10.16 & 7.91  & 41.90 & 16.73 & 34.17 \\
Bilateral filter             & 10.39 & 7.96  & 41.97 & 16.96 & 34.43 \\\midrule
No stereo method             & 30.84 & 28.33 & 59.57 & 64.53 & 39.23 \\
Full stereo method           & 9.87  & 7.91  & 41.86 & 16.72 & 33.89 \\\bottomrule
\end{tabular}
\end{center}
\caption{The numbers measure validation error when a particular
post-processing step is excluded from the stereo method.  The last two rows of
the tables should be interpreted differently: they contain the validation error
of the raw convolutional neural network and the validation error after the
complete stereo method.  For example, if we exclude semiglobal matching, the
fast architecture achieves an error rate of 8.78\,\% on the KITTI 2012 data
set and an error rate of 3.02\,\% after applying the full stereo method.
We abbreviate the method names as ``fst'' for the fast architecture, ``acrt''
for the accurate architecture, ``sad'' for the sum of absolute differences,
``cens'' for the census transform, and ``ncc'' for the normalized
cross-correlation matching cost. }

\label{tbl:sm_skip}
\end{table}

\subsection{Stereo Method}

The stereo method includes a number of post-processing steps: cross-based cost
aggregation, semi\-global matching, interpolation, subpixel enhancement, a
median, and a bilateral filter. We ran a set of experiments in which we excluded
each of the aforementioned steps and recorded the validation error (see
Table~\ref{tbl:sm_skip}).

The last two rows of Table~\ref{tbl:sm_skip} allude to the importance of the
post-processing steps of the stereo method. We see that, if all post-processing
steps are removed, the validation error of the accurate architecture increases
from 2.61\,\% to 13.49\,\% on KITTI 2012, from 3.25\,\% to 13.38\,\% on KITTI
2015, and from 7.91\,\% to 28.33\,\% on Middlebury. 

Out of all post-processing steps of the stereo method, semiglobal matching
affects the validation error the strongest. If we remove it, the validation
error increases from 2.61\,\% to 4.26\,\% on KITTI 2012, from 3.25\,\% to
4.51\,\% on KITTI 2015, and from 7.91\,\% to 11.99\,\% on Middlebury. 

We did not use the left-right consistency check to eliminate errors in occluded
regions on the Middlebury data set.  The error rate increased from 7.91\,\% to
8.22\,\% using the left-right consistency check on the accurate architecture,
which is why we decided to remove it.

\subsection{Data Set Size}

We used a supervised learning approach to measure the similarity between
image patches. It is, therefore, natural to ask how does the size of the data
set affect the quality of the disparity maps. To answer this question, we
retrain our networks on smaller training sets obtained by selecting a random
set of examples (see Table~\ref{tbl:dataset_size}).

\begin{table}[tb]
\begin{center}
\begin{tabular}{c ccc ccc}
\toprule
& \multicolumn{2}{c}{KITTI 2012} & \multicolumn{2}{c}{KITTI 2015} & \multicolumn{2}{c}{Middlebury} \\
\cmidrule(lr){2-3}
\cmidrule(lr){4-5}
\cmidrule(lr){6-7}
Data Set Size (\%) & fst & acrt & fst & acrt & fst & acrt\\\midrule
20 & 3.17 & 2.84 & 4.13 & 3.53 & 11.14 & 9.73 \\
40 & 3.11 & 2.75 & 4.10 & 3.40 & 10.35 & 8.71 \\
60 & 3.09 & 2.67 & 4.05 & 3.34 & 10.14 & 8.36 \\
80 & 3.05 & 2.65 & 4.02 & 3.29 & 10.09 & 8.21 \\
100 & 3.02 & 2.61 & 3.99 & 3.25 & 9.87 & 7.91 \\
\bottomrule
\end{tabular}
\caption{The validation error as a function of training set size.}

\label{tbl:dataset_size}
\end{center}
\end{table}

We observe that the validation error decreases as we increase the number of
training examples. These experiments suggest a simple strategy for improving the
results of our stereo method: collect a larger data set.

\subsection{Transfer Learning}

\begin{table}[tb]
\begin{center}
\begin{tabular}{cc cccccc}
\toprule
&& \multicolumn{6}{c}{Test Set} \\\cmidrule(lr){3-8}
&& \multicolumn{2}{c}{KITTI 2012} & \multicolumn{2}{c}{KITTI 2015} & \multicolumn{2}{c}{Middlebury} \\
\cmidrule(lr){3-4}
\cmidrule(lr){5-6}
\cmidrule(lr){7-8}
&& fst & acrt & fst & acrt & fst & acrt\\\midrule
\multirow{3}{*}{Training Set} 
& KITTI 2012 & 3.02 & 2.61 & 4.12 & 3.99 & 12.78 & 11.09 \\
& KITTI 2015 & 3.60 & 4.28 & 3.99 & 3.25 & 13.70 & 14.19 \\
& Middlebury & 3.16 & 3.07 & 4.48 & 4.49 & 9.87 & 7.91  \\
\bottomrule
\end{tabular}
\caption{The validation error when the training and test sets differ. For
example, the validation error is 3.16\,\% when the Middlebury data set is used
for training the fast architecture and the trained network is tested on the
KITTI 2012 data set.}

\label{tbl:transfer}
\end{center}
\end{table}

Up to this point the training and validation sets were created from the same
stereo data set, either KITTI 2012, KITTI 2015, or Middlebury. To evaluate the
performance of our method in the transfer learning setting, we run experiments
where the validation error is computed on a different data set than the one
used for training. For example, we would use the Middlebury data set to train
the matching cost neural network and evaluate its performance on the KITTI 2012
data set. These experiments give us some idea of the expected performance in a
real-world application, where it isn't possible to train a specialized network
because no ground truth is available. The results of these experiments are
shown in Table~\ref{tbl:transfer}.

Some results in Table~\ref{tbl:transfer} were unexpected. For example,
the validation error on KITTI 2012 is lower when using the Middlebury training
set compared to the KITTI 2015 training set, even though the KITTI 2012 data
set is obviously more similar to KITTI 2015 than Middlebury. Furthermore, the
validation error on KITTI 2012 is lower when using the fast architecture
instead of the accurate architecture when training on KITTI 2015.

The matching cost neural network trained on the Middlebury data set transfers
well to the KITTI data sets. Its validation error is similar to the validation
errors obtained by networks trained on the KITTI data sets.

\subsection{Hyperparameters}

\begin{table}[p]
\begin{center}
\begin{tabular}{lc ccc ccc}
\toprule
& & \multicolumn{2}{c}{KITTI 2012} & \multicolumn{2}{c}{KITTI 2015} & \multicolumn{2}{c}{Middlebury} \\
\cmidrule(lr){3-4}
\cmidrule(lr){5-6}
\cmidrule(lr){7-8}
Hyperparameter & & fst & acrt & fst & acrt & fst & acrt\\\midrule

\multirow{6}{*}{\texttt{num\_conv\_layers}}
& 1 & 5.96 & 3.97 & 5.61 & 4.06 & 20.74 & 12.37 \\
& 2 & 3.52 & 2.98 & 4.19 & 3.45 & 12.11 & 9.20 \\
& 3 & 3.10 & 2.72 & 4.04 & 3.27 & 10.81 & 8.56 \\
& 4 & 3.02 & 2.61 & 3.99 & 3.25 & 10.26 & 8.21 \\
& 5 & 3.03 & 2.64 & 3.99 & 3.30 & 9.87 & 7.91 \\
& 6 & 3.05 & 2.70 & 4.01 & 3.38 & 9.71 & 8.11 \\
\midrule
\multirow{8}{*}{\texttt{num\_conv\_feature\_maps}}
& 16 & 3.33 & 2.84 & 4.32 & 3.51 & 11.79 & 10.06 \\
& 32 & 3.15 & 2.68 & 4.12 & 3.35 & 10.48 & 8.67 \\
& 48 & 3.07 & 2.66 & 4.06 & 3.32 & 10.20 & 8.47 \\
& 64 & 3.02 & 2.64 & 3.99 & 3.30 & 9.87 & 8.12 \\
& 80 & 3.02 & 2.64 & 3.99 & 3.29 & 9.81 & 7.95 \\
& 96 & 2.99 & 2.68 & 3.97 & 3.27 & 9.62 & 8.03 \\
& 112 & 2.98 & 2.61 & 3.96 & 3.25 & 9.59 & 7.91 \\
& 128 & 2.97 & 2.63 & 3.95 & 3.23 & 9.45 & 7.92 \\\midrule
\multirow{5}{*}{\texttt{num\_fc\_layers}}
& 1 & & 2.83 & & 3.50 & & 8.52 \\
& 2 & & 2.70 & & 3.31 & & 8.33 \\
& 3 & & 2.62 & & 3.30 & & 8.06 \\
& 4 & & 2.61 & & 3.25 & & 8.00 \\
& 5 & & 2.62 & & 3.29 & & 7.91 \\\midrule
\multirow{4}{*}{\texttt{num\_fc\_units}}
& 128 & & 2.72 & & 3.36 & & 8.44 \\
& 256 & & 2.65 & & 3.28 & & 8.03 \\
& 384 & & 2.61 & & 3.25 & & 7.91 \\
& 512 & & 2.60 & & 3.23 & & 7.90 \\\midrule
\multirow{5}{*}{\texttt{dataset\_neg\_low}}
& 1.0 & 3.00 & 2.76 & 3.97 & 3.35 & 9.84 & 8.00 \\
& 1.5 & 3.00 & 2.71 & 3.97 & 3.33 & 9.87 & 7.91 \\
& 2.0 & 2.99 & 2.63 & 3.98 & 3.31 & 9.98 & 8.08 \\
& 4.0 & 3.02 & 2.61 & 3.99 & 3.25 & 10.20 & 8.66 \\
& 6.0 & 3.06 & 2.63 & 4.05 & 3.28 & 10.13 & 8.86 \\\midrule
\multirow{5}{*}{\texttt{dataset\_neg\_high}}
& 6 & 3.00 & 2.72 & 3.98 & 3.30 & 9.87 & 8.59 \\
& 10 & 3.02 & 2.61 & 3.99 & 3.25 & 9.97 & 8.23 \\
& 14 & 3.04 & 2.61 & 4.02 & 3.25 & 10.00 & 8.05 \\
& 18 & 3.07 & 2.60 & 4.06 & 3.23 & 9.98 & 8.11 \\
& 22 & 3.07 & 2.61 & 4.05 & 3.24 & 10.16 & 7.91 \\\midrule
\multirow{5}{*}{\texttt{dataset\_pos}}
& 0.0 & 3.04 & 2.67 & 4.00 & 3.26 & 9.92 & 7.97 \\
& 0.5 & 3.02 & 2.65 & 3.99 & 3.28 & 9.87 & 7.91 \\
& 1.0 & 3.02 & 2.61 & 3.99 & 3.25 & 9.86 & 8.04 \\
& 1.5 & 3.04 & 2.62 & 4.04 & 3.27 & 10.00 & 8.34 \\
& 2.0 & 3.04 & 2.66 & 4.04 & 3.29 & 10.16 & 8.51 \\\bottomrule
\end{tabular}
\caption{Validation errors computed across a range of hyperparameter settings.}

\label{tbl:hyperparameters}
\end{center}
\end{table}

Searching for a good set of hyperparameters is a daunting task---with the
search space growing exponentially with the number of hyperparameters and no
gradient to guide us. To better understand the effect of each hyperparameter on
the validation error, we conduct a series of experiments where we vary the
value of one hyperparameter while keeping the others fixed to their
default values. The results are shown in Table~\ref{tbl:hyperparameters} and
can be summarized by observing that increasing the size of the network improves
the generalization performance, but only up to a point, when presumably,
because of the size of the data set, the generalization performance starts do
decrease.  

Note that the \texttt{num\_conv\_layers} hyperparameter implicitly controls
the size of the image patches. For example, a network with one convolutional
layer with \(3 \times 3\) kernels compares image patches of size \(3 \times
3\), while a network with five convolutional layers compares patches of size
\(11 \times 11\). 

\section{Conclusion}

We presented two convolutional neural network architectures for learning a
similarity measure on image patches and applied them to the problem of stereo
matching. 

The source code of our implementation is available at
\url{https://github.com/jzbontar/mc-cnn}.  The online repository contains
procedures for computing the disparity map, training the network, as well as
the post-processing steps of the stereo method.

The accurate architecture produces disparity maps with lower error rates than
any previously published method on the KITTI 2012, KITTI 2015, and Middlebury
data sets. The fast architecture computes the disparity maps up to 90 times
faster than the accurate architecture with only a small increase in error.
These results suggest that convolutional neural networks are well suited for
computing the stereo matching cost even for applications that require real-time
performance.

The fact that a relatively simple convolutional neural network outperformed all
previous methods on the well-studied problem of stereo is a rather important
demonstration of the power of modern machine learning approaches.

\bibliography{paper}

\begin{thebibliography}{50}
\providecommand{\natexlab}[1]{#1}
\providecommand{\url}[1]{\texttt{#1}}
\expandafter\ifx\csname urlstyle\endcsname\relax
  \providecommand{\doi}[1]{doi: #1}\else
  \providecommand{\doi}{doi: \begingroup \urlstyle{rm}\Url}\fi

\bibitem[Bradski(2000)]{opencv_library}
G.~Bradski.
\newblock The {OpenCV} library.
\newblock \emph{Dr. Dobb's Journal of Software Tools}, 2000.

\bibitem[Bromley et~al.(1993)Bromley, Bentz, Bottou, Guyon, LeCun, Moore,
  S{\"a}ckinger, and Shah]{bromley1993signature}
Jane Bromley, James~W Bentz, L{\'e}on Bottou, Isabelle Guyon, Yann LeCun, Cliff
  Moore, Eduard S{\"a}ckinger, and Roopak Shah.
\newblock Signature verification using a siamese time delay neural network.
\newblock \emph{International Journal of Pattern Recognition and Artificial
  Intelligence}, 7\penalty0 (04):\penalty0 669--688, 1993.

\bibitem[Brown et~al.(2011)Brown, Hua, and Winder]{brown2011discriminative}
Matthew Brown, Gang Hua, and Simon Winder.
\newblock Discriminative learning of local image descriptors.
\newblock \emph{IEEE Transactions on Pattern Analysis and Machine
  Intelligence}, 33\penalty0 (1):\penalty0 43--57, 2011.

\bibitem[Brox and Malik(2011)]{brox2011large}
Thomas Brox and Jitendra Malik.
\newblock Large displacement optical flow: descriptor matching in variational
  motion estimation.
\newblock \emph{IEEE Transactions on Pattern Analysis and Machine
  Intelligence}, 33\penalty0 (3):\penalty0 500--513, 2011.

\bibitem[Chakrabarti et~al.(2015)Chakrabarti, Xiong, Gortler, and
  Zickler]{chakrabarti2014low}
Ayan Chakrabarti, Ying Xiong, Steven~J. Gortler, and Todd Zickler.
\newblock Low-level vision by consensus in a spatial hierarchy of regions.
\newblock \emph{IEEE Conference on Computer Vision and Pattern Recognition
  (CVPR)}, June 2015.

\bibitem[Chen et~al.(2015)Chen, Sun, Yu, Wang, and Huang]{deep_embed}
Zhuoyuan Chen, Xun Sun, Yinan Yu, Liang Wang, and Chang Huang.
\newblock A deep visual correspondence embedding model for stereo matching
  costs.
\newblock \emph{IEEE International Conference on Computer Vision (ICCV)}, 2015.

\bibitem[Chetlur et~al.(2014)Chetlur, Woolley, Vandermersch, Cohen, Tran,
  Catanzaro, and Shelhamer]{chetlur2014cudnn}
Sharan Chetlur, Cliff Woolley, Philippe Vandermersch, Jonathan Cohen, John
  Tran, Bryan Catanzaro, and Evan Shelhamer.
\newblock {cuDNN}: Efficient primitives for deep learning.
\newblock \emph{CoRR}, abs/1410.0759, 2014.
\newblock URL \url{http://arxiv.org/abs/1410.0759}.

\bibitem[Collobert et~al.(2011)Collobert, Kavukcuoglu, and
  Farabet]{collobert2011torch7}
Ronan Collobert, Koray Kavukcuoglu, and Cl{\'e}ment Farabet.
\newblock Torch7: A matlab-like environment for machine learning.
\newblock In \emph{BigLearn, NIPS Workshop}, 2011.

\bibitem[Einecke and Eggert(2010)]{einecke2010two}
Nils Einecke and Julian Eggert.
\newblock A two-stage correlation method for stereoscopic depth estimation.
\newblock In \emph{Digital Image Computing: International Conference on
  Techniques and Applications (DICTA)}, pages 227--234, 2010.

\bibitem[Geiger et~al.(2011)Geiger, Roser, and Urtasun]{geiger2011efficient}
Andreas Geiger, Martin Roser, and Raquel Urtasun.
\newblock Efficient large-scale stereo matching.
\newblock In \emph{Proceedings of the 10th Asian Conference on Computer Vision
  - Volume Part I}, ACCV'10, pages 25--38. Springer-Verlag, Berlin, Heidelberg,
  2011.

\bibitem[Geiger et~al.(2013)Geiger, Lenz, Stiller, and Urtasun]{Geiger2013IJRR}
Andreas Geiger, Philip Lenz, Christoph Stiller, and Raquel Urtasun.
\newblock Vision meets robotics: the {KITTI} dataset.
\newblock \emph{International Journal of Robotics Research (IJRR)}, 2013.

\bibitem[G{\"u}ney and Geiger(2015)]{guney2015displets}
Fatma G{\"u}ney and Andreas Geiger.
\newblock Displets: Resolving stereo ambiguities using object knowledge.
\newblock \emph{IEEE Conference on Computer Vision and Pattern Recognition
  (CVPR)}, June 2015.

\bibitem[Haeusler et~al.(2013)Haeusler, Nair, and
  Kondermann]{haeusler2013ensemble}
Ralf Haeusler, Rahul Nair, and Daniel Kondermann.
\newblock Ensemble learning for confidence measures in stereo vision.
\newblock \emph{IEEE Conference on Computer Vision and Pattern Recognition
  (CVPR)}, June 2013.

\bibitem[Han et~al.(2015)Han, Leung, Jia, Sukthankar, and
  Berg]{han2015matchnet}
Xufeng Han, Thomas Leung, Yangqing Jia, Rahul Sukthankar, and Alexander~C Berg.
\newblock {MatchNet}: Unifying feature and metric learning for patch-based
  matching.
\newblock \emph{IEEE Conference on Computer Vision and Pattern Recognition
  (CVPR)}, June 2015.

\bibitem[Hirschm{\"u}ller(2008)]{hirschmuller2008stereo}
Heiko Hirschm{\"u}ller.
\newblock Stereo processing by semiglobal matching and mutual information.
\newblock \emph{IEEE Transactions on Pattern Analysis and Machine
  Intelligence}, 30\penalty0 (2):\penalty0 328--341, 2008.

\bibitem[Hirschm{\"u}ller and Scharstein(2007)]{hirschmuller2007evaluation}
Heiko Hirschm{\"u}ller and Daniel Scharstein.
\newblock Evaluation of cost functions for stereo matching.
\newblock \emph{IEEE Conference on Computer Vision and Pattern Recognition
  (CVPR)}, 2007.

\bibitem[Hirschm{\"u}ller and Scharstein(2009)]{hirschmuller2009evaluation}
Heiko Hirschm{\"u}ller and Daniel Scharstein.
\newblock Evaluation of stereo matching costs on images with radiometric
  differences.
\newblock \emph{IEEE Transactions on Pattern Analysis and Machine
  Intelligence}, 31\penalty0 (9):\penalty0 1582--1599, 2009.

\bibitem[Hornacek et~al.(2014)Hornacek, Fitzgibbon, and
  Rother]{hornacek2014sphereflow}
Michael Hornacek, Andrew Fitzgibbon, and Carsten Rother.
\newblock {SphereFlow}: 6 {DoF} scene flow from {RGB-D} pairs.
\newblock \emph{IEEE Conference on Computer Vision and Pattern Recognition
  (CVPR)}, June 2014.

\bibitem[Kong and Tao(2004)]{kong2004method}
Dan Kong and Hai Tao.
\newblock A method for learning matching errors for stereo computation.
\newblock \emph{British Machine Vision Conference (BMVC)}, 2004.

\bibitem[Kong and Tao(2006)]{kong2006stereo}
Dan Kong and Hai Tao.
\newblock Stereo matching via learning multiple experts behaviors.
\newblock \emph{British Machine Vision Conference (BMVC)}, 2006.

\bibitem[Kostkov{\'a} and S{\'a}ra(2003)]{kostkova2003stratified}
Jana Kostkov{\'a} and Radim S{\'a}ra.
\newblock Stratified dense matching for stereopsis in complex scenes.
\newblock \emph{British Machine Vision Conference (BMVC)}, 2003.

\bibitem[Kowalczuk et~al.(2013)Kowalczuk, Psota, and Perez]{kowalczuk2013real}
Jedrzej Kowalczuk, Eric~T Psota, and Lance~C Perez.
\newblock Real-time stereo matching on {CUDA} using an iterative refinement
  method for adaptive support-weight correspondences.
\newblock \emph{IEEE Transactions on Circuits and Systems for Video
  Technology}, 23\penalty0 (1):\penalty0 94--104, 2013.

\bibitem[LeCun et~al.(1998)LeCun, Bottou, Bengio, and
  Haffner]{lecun1998gradient}
Yann LeCun, L{\'e}on Bottou, Yoshua Bengio, and Patrick Haffner.
\newblock Gradient-based learning applied to document recognition.
\newblock \emph{Proceedings of the IEEE}, 86\penalty0 (11):\penalty0
  2278--2324, 1998.

\bibitem[Li and Huttenlocher(2008)]{li2008learning}
Yunpeng Li and Daniel~P Huttenlocher.
\newblock Learning for stereo vision using the structured support vector
  machine.
\newblock \emph{IEEE Conference on Computer Vision and Pattern Recognition
  (CVPR)}, June 2008.

\bibitem[Mei et~al.(2011)Mei, Sun, Zhou, Wang, Zhang, et~al.]{mei2011building}
Xing Mei, Xun Sun, Mingcai Zhou, Haitao Wang, Xiaopeng Zhang, et~al.
\newblock On building an accurate stereo matching system on graphics hardware.
\newblock \emph{IEEE International Conference on Computer Vision Workshops
  (ICCV Workshops)}, pages 467--474, 2011.

\bibitem[Menze and Geiger(2015)]{menze2015object}
Moritz Menze and Andreas Geiger.
\newblock Object scene flow for autonomous vehicles.
\newblock \emph{IEEE Conference on Computer Vision and Pattern Recognition
  (CVPR)}, June 2015.

\bibitem[Nickolls et~al.(2008)Nickolls, Buck, Garland, and
  Skadron]{nickolls2008scalable}
John Nickolls, Ian Buck, Michael Garland, and Kevin Skadron.
\newblock Scalable parallel programming with {CUDA}.
\newblock \emph{Queue}, 6\penalty0 (2):\penalty0 40--53, 2008.

\bibitem[Paulin et~al.(2015)Paulin, Douze, Harchaoui, Mairal, Perronin, and
  Schmid]{paulin2015local}
Mattis Paulin, Matthijs Douze, Zaid Harchaoui, Julien Mairal, Florent Perronin,
  and Cordelia Schmid.
\newblock Local convolutional features with unsupervised training for image
  retrieval.
\newblock In \emph{IEEE International Conference on Computer Vision (ICCV)},
  pages 91--99, 2015.

\bibitem[Peris et~al.(2012)Peris, Maki, Martull, Ohkawa, and
  Fukui]{peris2012towards}
Martin Peris, Atsuto Maki, Sara Martull, Yasuhiro Ohkawa, and Kazuhiro Fukui.
\newblock Towards a simulation driven stereo vision system.
\newblock In \emph{21st International Conference on Pattern Recognition
  (ICPR)}, pages 1038--1042, 2012.

\bibitem[Psota et~al.(2015)Psota, Kowalczuk, Mittek, and Perez]{psota2015map}
Eric~T Psota, Jedrzej Kowalczuk, Mateusz Mittek, and Lance~C Perez.
\newblock Map disparity estimation using hidden markov trees.
\newblock \emph{IEEE International Conference on Computer Vision (ICCV)}, 2015.

\bibitem[Revaud et~al.(2015)Revaud, Weinzaepfel, Harchaoui, and
  Schmid]{revaud2015deepmatching}
Jerome Revaud, Philippe Weinzaepfel, Zaid Harchaoui, and Cordelia Schmid.
\newblock Deepmatching: Hierarchical deformable dense matching.
\newblock \emph{ArXiv e-prints}, 1\penalty0 (7):\penalty0 8, 2015.

\bibitem[Scharstein and Pal(2007)]{scharstein2007learning}
Daniel Scharstein and Chris Pal.
\newblock Learning conditional random fields for stereo.
\newblock \emph{IEEE Conference on Computer Vision and Pattern Recognition
  (CVPR)}, June 2007.

\bibitem[Scharstein and Szeliski(2002)]{scharstein2002taxonomy}
Daniel Scharstein and Richard Szeliski.
\newblock A taxonomy and evaluation of dense two-frame stereo correspondence
  algorithms.
\newblock \emph{International Journal of Computer Vision}, 47\penalty0
  (1-3):\penalty0 7--42, 2002.

\bibitem[Scharstein and Szeliski(2003)]{scharstein2003high}
Daniel Scharstein and Richard Szeliski.
\newblock High-accuracy stereo depth maps using structured light.
\newblock \emph{IEEE Conference on Computer Vision and Pattern Recognition
  (CVPR)}, June 2003.

\bibitem[Scharstein et~al.(2014)Scharstein, Hirschm{\"u}ller, Kitajima,
  Krathwohl, Ne{\v{s}}i{\'c}, Wang, and Westling]{scharstein2014high}
Daniel Scharstein, Heiko Hirschm{\"u}ller, York Kitajima, Greg Krathwohl, Nera
  Ne{\v{s}}i{\'c}, Xi~Wang, and Porter Westling.
\newblock High-resolution stereo datasets with subpixel-accurate ground truth.
\newblock \emph{German Conference on Pattern Recognition (GCPR)}, September
  2014.

\bibitem[Simonyan et~al.(2014)Simonyan, Vedaldi, and
  Zisserman]{simonyan2014learning}
Karen Simonyan, Andrea Vedaldi, and Andrew Zisserman.
\newblock Learning local feature descriptors using convex optimisation.
\newblock \emph{IEEE Transactions on Pattern Analysis and Machine
  Intelligence}, 36\penalty0 (8):\penalty0 1573--1585, 2014.

\bibitem[Sinha et~al.(2014)Sinha, Scharstein, and Szeliski]{sinha2014efficient}
Sudipta~N Sinha, Daniel Scharstein, and Richard Szeliski.
\newblock Efficient high-resolution stereo matching using local plane sweeps.
\newblock \emph{IEEE Conference on Computer Vision and Pattern Recognition
  (CVPR)}, June 2014.

\bibitem[Spyropoulos et~al.(2014)Spyropoulos, Komodakis, and
  Mordohai]{spyropoulos2014learning}
Aristotle Spyropoulos, Nikos Komodakis, and Philippos Mordohai.
\newblock Learning to detect ground control points for improving the accuracy
  of stereo matching.
\newblock \emph{IEEE Conference on Computer Vision and Pattern Recognition
  (CVPR)}, June 2014.

\bibitem[Sun et~al.(2014)Sun, Roth, and Black]{sun2014quantitative}
Deqing Sun, Stefan Roth, and Michael~J Black.
\newblock A quantitative analysis of current practices in optical flow
  estimation and the principles behind them.
\newblock \emph{International Journal of Computer Vision}, 106\penalty0
  (2):\penalty0 115--137, 2014.

\bibitem[Trzcinski et~al.(2012)Trzcinski, Christoudias, Lepetit, and
  Fua]{trzcinski2012learning}
Tomasz Trzcinski, Mario Christoudias, Vincent Lepetit, and Pascal Fua.
\newblock Learning image descriptors with the boosting-trick.
\newblock In \emph{Advances in neural information processing systems}, pages
  269--277, 2012.

\bibitem[Vogel et~al.(2013)Vogel, Schindler, and Roth]{vogel2013piecewise}
Christoph Vogel, Konrad Schindler, and Stefan Roth.
\newblock Piecewise rigid scene flow.
\newblock \emph{IEEE International Conference on Computer Vision (ICCV)}, 2013.

\bibitem[Vogel et~al.(2014)Vogel, Roth, and Schindler]{vogel2014view}
Christoph Vogel, Stefan Roth, and Konrad Schindler.
\newblock View-consistent {3D} scene flow estimation over multiple frames.
\newblock \emph{European Conference on Computer Vision (ECCV)}, September 2014.

\bibitem[Vogel et~al.(2015)Vogel, Schindler, and Roth]{vogel20153d}
Christoph Vogel, Konrad Schindler, and Stefan Roth.
\newblock {3D} scene flow estimation with a piecewise rigid scene model.
\newblock \emph{International Journal of Computer Vision}, pages 1--28, 2015.

\bibitem[Yamaguchi et~al.(2014)Yamaguchi, McAllester, and
  Urtasun]{yamaguchi2014efficient}
Koichiro Yamaguchi, David McAllester, and Raquel Urtasun.
\newblock Efficient joint segmentation, occlusion labeling, stereo and flow
  estimation.
\newblock \emph{European Conference on Computer Vision (ECCV)}, September 2014.

\bibitem[Zabih and Woodfill(1994)]{zabih1994non}
Ramin Zabih and John Woodfill.
\newblock Non-parametric local transforms for computing visual correspondence.
\newblock \emph{European Conference on Computer Vision (ECCV)}, 1994.

\bibitem[Zagoruyko and Komodakis(2015)]{zagoruyko2015learning}
Sergey Zagoruyko and Nikos Komodakis.
\newblock Learning to compare image patches via convolutional neural networks.
\newblock \emph{IEEE Conference on Computer Vision and Pattern Recognition
  (CVPR)}, June 2015.

\bibitem[{\v{Z}}bontar and LeCun(2015)]{Zbontar_2015_CVPR}
Jure {\v{Z}}bontar and Yann LeCun.
\newblock Computing the stereo matching cost with a convolutional neural
  network.
\newblock \emph{IEEE Conference on Computer Vision and Pattern Recognition
  (CVPR)}, June 2015.

\bibitem[Zhang et~al.(2015)Zhang, Li, Cheng, Cai, Chao, and
  Rui]{zhang2015meshstereo}
Chi Zhang, Zhiwei Li, Yanhua Cheng, Rui Cai, Hongyang Chao, and Yong Rui.
\newblock Meshstereo: A global stereo model with mesh alignment regularization
  for view interpolation.
\newblock \emph{IEEE International Conference on Computer Vision (ICCV)}, 2015.

\bibitem[Zhang et~al.(2009)Zhang, Lu, and Lafruit]{zhang2009cross}
Ke~Zhang, Jiangbo Lu, and Gauthier Lafruit.
\newblock Cross-based local stereo matching using orthogonal integral images.
\newblock \emph{IEEE Transactions on Circuits and Systems for Video
  Technology}, 19\penalty0 (7):\penalty0 1073--1079, 2009.

\bibitem[Zhang and Seitz(2007)]{zhang2007estimating}
Li~Zhang and Steven~M Seitz.
\newblock Estimating optimal parameters for {MRF} stereo from a single image
  pair.
\newblock \emph{IEEE Transactions on Pattern Analysis and Machine
  Intelligence}, 29\penalty0 (2):\penalty0 331--342, 2007.

\end{thebibliography}

\end{document}